\pdfoutput=1
\documentclass[11pt]{article}

\usepackage[]{latex/acl}
\usepackage{tabularx, booktabs}
\usepackage{arydshln} % 引入宏包
\usepackage{times}
\usepackage{soul}
\usepackage{listings}
\usepackage{latexsym}
\usepackage{makecell}
\usepackage{graphicx}
\usepackage{multirow}
\usepackage{amsmath} % 用于数学符号
\usepackage{amssymb} % 额外的数学符号
\usepackage{CJKutf8}
\usepackage{paralist}

\usepackage[T1]{fontenc}
\usepackage[utf8]{inputenc}

\usepackage{microtype}
\usepackage{inconsolata}
\usepackage{cleveref}
\usepackage{pifont}
\title{Self-Contrast: Better Reflection Through Inconsistent Solving Perspectives}

\author{Wenqi Zhang$^{1}$, Yongliang Shen$^{1}$, Linjuan Wu$^{1}$\\
       {\bf Qiuying Peng$^{2}$, Jun Wang$^{2}$, Yueting Zhuang$^{1}$, Weiming Lu$^{1 \dagger}$}\\
  $^1$College of Computer Science and Technology, Zhejiang University \\
  $^2$OPPO Research Institute, China \\
  \texttt{\{zhangwenqi, luwm\}@zju.edu.cn}} 

\begin{document}
\maketitle
\renewcommand{\thefootnote}{\fnsymbol{footnote}} %将脚注符号设置为fnsymbol类型，即特殊符号表示
\footnotetext[2]{Corresponding author.}  %对应脚注[1]
\renewcommand{\thefootnote}{\arabic{footnote}}%将脚注符号重新改和数
\begin{abstract}
The reflection capacity of Large Language Model (LLM) has garnered extensive attention. A post-hoc prompting strategy, e.g., reflexion and self-refine, refines LLM's response based on self-evaluated or external feedback. However, recent research indicates without external feedback, LLM's intrinsic reflection is unstable. Our investigation unveils that the key bottleneck is the quality of the self-evaluated feedback. We find LLMs often exhibit overconfidence or high randomness when self-evaluate, offering stubborn or inconsistent feedback, which causes poor reflection. To remedy this, we advocate \textbf{Self-Contrast}: It adaptively \textbf{explores} diverse solving perspectives tailored to the request, \textbf{contrasts} the differences, and \textbf{summarizes} these discrepancies into a checklist which could be used to re-examine and eliminate discrepancies. Our method provides LLM with diverse perspectives to alleviate stubborn biases. Moreover, their discrepancies indicate potential errors or inherent uncertainties that LLM often overlooks. Reflecting upon these can prompt more accurate and stable reflection. Experiments conducted on a series of reasoning and translation tasks with different LLMs serve to underscore the effectiveness and generality of our strategy.

%catalyze more accurate and stable reflection  since these discrepancies often indicate potential errors or inherent uncertainties that LLM often overlooks. 

%Reflecting upon these can catalyze more accurate and stable reflection.

%Empirically, we demonstrate that Self-Contrast performs well in reasoning and translation tasks with reasonable invocation cost. 

%for accurate and stable reflection

%Therefore, we instruct LLM to reflect on the reason causing discrepancies and summarize them into multiple re-examining instructions to resolve discrepancies and achieve more accurate and stable reflection. 

%LLMs consider \textbf{where} the difference is, \textbf{why} different, and \textbf{how} to solve.
%Concretely, Self-Contrast first adaptively creates multiple perspectives tailored to the question and organizes them in a relational graph, where each result is a vertex, and their discrepancies act as edges. Then reflection is performed by mitigating the disparities on each edge. 

%In light of this, we are curious: \textbf{Can LLMs harvest insights by contrasting the difference between these candidates?} 

%Therefore, we advocate Self-Contrast: LLMs adaptively devise multiple solving perspectives and autonomously contrast their inconsistency for reflection. Rather than reflection within a single perspective, 

\end{abstract}
%Rather than reflection within a single perspective, our relation graph empowers LLMs to group, aggregate, and modify inconsistent perspectives for better reflection. for reflection after node grouping, aggregating, and contrasting.
%Compared to vanilla reflection, we achieve improvements of +8\% and +10\% on the GSM8K and SVAMP. It suggests contrasts between inconsistent perspectives may act as "external" feedback to overcome LLM's stubborn biases and reduce indecision.

\section{Introduction}
Mastering reasoning and decision-making capabilities is of utmost importance to paving the way for artificial general intelligence. Recently, large language models (LLMs) \citep{brown2020language, chowdhery2022palm, zhang2022opt, Zeng2023GLM, touvron2023llama, Chatgpt, gpt4, Touvron2023Llama2O} and applications built on them \citep{schick2023toolformer, wu2023visual, shen2023hugginggpt, Zhang2023DataCopilotBB} demonstrate impressive capabilities in various domains, especially combined with Chain-of-Thought~\citep{wei2022chain, kojima2022large}, ReAct~\citep{Yao2022ReActSR}, Tree-of-Thought~\citep{Yao2023TreeOT} and other prompting techniques~\citep{gao2022pal, Wang2023Self, Zhou2022LeasttoMostPE, besta2023graph}.

\begin{figure}[!t] %H为当前位置，!htb为忽略美学标准，htbp为浮动图形
\centering %图片居中
\includegraphics[width=0.5\textwidth]{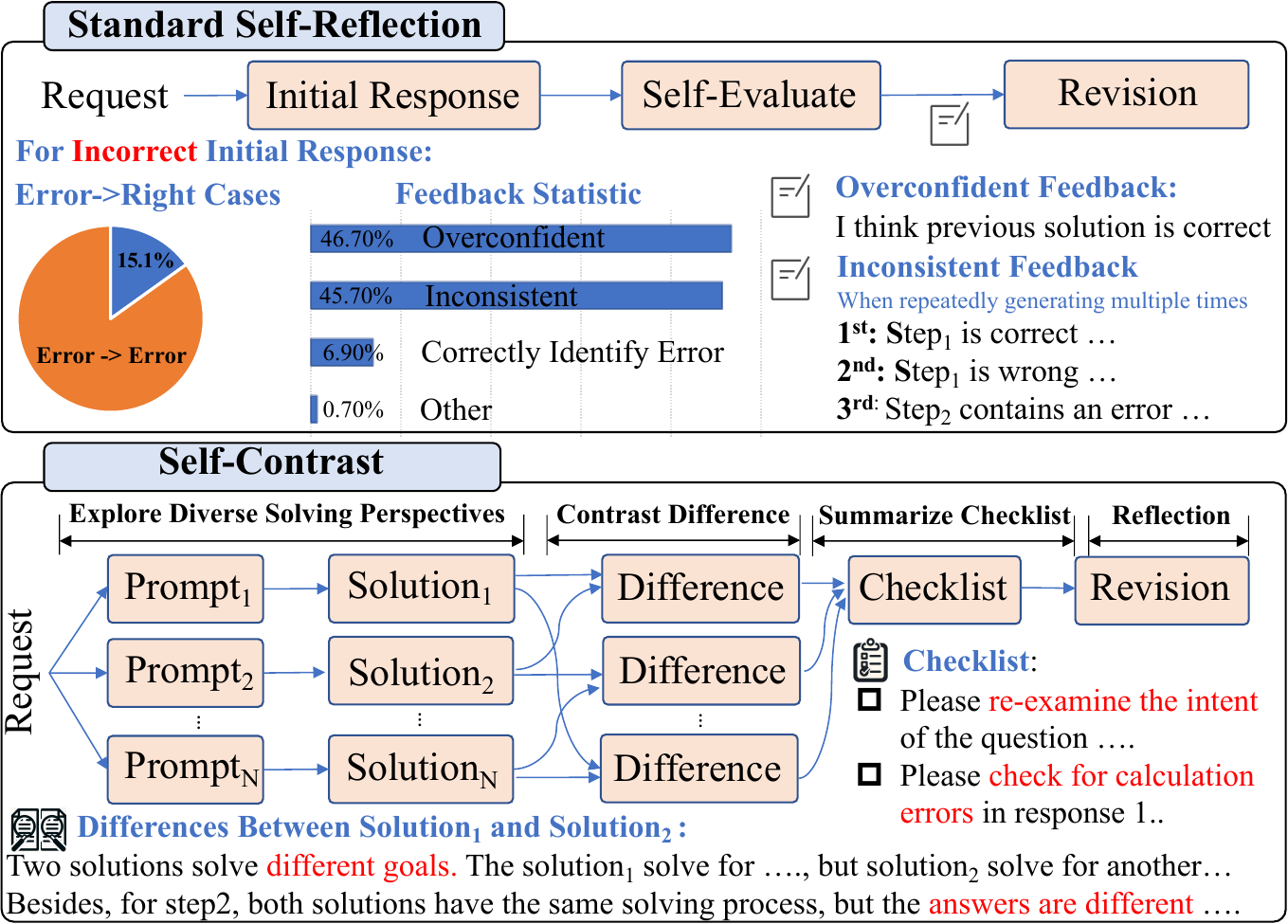} %
\caption{LLMs evaluate the initial response and provide feedback for revision. However, most erroneous responses remain uncorrected after reflection as the feedback is either overconfident (46.7\%) or inconsistent (45.7\%). Bottom: Self-Contrast \color{black}{explores} \color{black}multiple solving perspectives, and \color{black}{contrast} \color{black} their differences, and \color{black}{summarize} \color{black}them into insightful checklist for self-correction.}\label{figure1} %最终文档中希望显示的图片标题 
\end{figure}

Despite these advancements, LLMs are not entirely reliable~\citep{Zheng2023WhyDC, Frieder2023MathematicalCO, Yuan2023HowWD} since they frequently produce inaccuracies results, such as misunderstanding a key concept, overlooking some crucial details. A post-hoc prompting strategy, e.g., self-reflection, garnered considerable attention~\citep{Shinn2023ReflexionAA, Madaan2023SelfRefineIR, Paul2023REFINERRF}. It first generates an initial response (Initial Response), then gathers external feedback or self-evaluated feedback (Evaluation Phase) to refine prior response (Revision)~\citep{Welleck2022GeneratingSB, Kadavath2022LanguageM, Chen2023TeachingLL, Liang2023EncouragingDT, Kim2023LanguageMC, Zheng2023ProgressiveHintPI, Du2023ImprovingFA, Xi2023SelfPolishER, Ganguli2023TheCF, Pan2023AutomaticallyCL}. Numerous studies proclaim this three-stage strategy (Initial Response$\rightarrow$Evaluation$\rightarrow$Revision), can endow LLMs with the potential to self-correct previous imperfect responses. For a time, this belief appeared to dominate the community.

%A substantial body of research is dedicated to mitigating the error in their previous reasoning path by post-hoc correction prompting strategy, e.g., self-reflection ~\citep{Shinn2023ReflexionAA}, self-refine~\citep{Madaan2023SelfRefineIR, Paul2023REFINERRF}, self-correct~\citep{Welleck2022GeneratingSB}, self-debug~\citep{Chen2023TeachingLL} and multi-agent debate~\citep{Liang2023EncouragingDT} and most contemporaneous works~\citep{Kim2023LanguageMC, Zheng2023ProgressiveHintPI, Du2023ImprovingFA, Xi2023SelfPolishER, Ganguli2023TheCF, Pan2023AutomaticallyCL}. 

%Similar to the process of introspection and examination in human cognition, this post-hoc strategy empowers LLMs to inspect whether the previous reasoning process is correct or not, provide feedback themselves, and then deduce an improved path. 
%This phenomenon is particularly egregious in scenarios requiring complex reasoning.

However, recent studies~\citep{Huang2023LargeLM, Stechly2023GPT4DK, Liang2023EncouragingDT, Valmeekam2023CanLL} have cast doubt on LLM's inherent reflection capability. Their research indicates that without external feedback, LLMs have difficulties in amending prior responses. It implies self-correction is unreliable when relying only on LLM itself and simple post-hoc prompting strategies.

%However, in the absence of an external "oracle" label, LLMs appear to struggle with correcting prior answers relying solely on inherent capabilities, sometimes even resulting in performance degradation.

%They argue that in the absence of external feedback, such post-hoc correction prompting is not universally effective. LLM exhibits considerable uncertainty when attempting to rectify errors in previous responses on its own, sometimes resulting in performance degradation. 

We are also intrigued by LLM's internal reflection ability, as external feedback is not available in most scenarios.
% Exploring how to rely on LLM itself for reliable reflection is compelling.
% Firstly, we conduct experiments to comprehensively investigate performance improvement caused by reflection, across multiple tasks, different prompts, and LLMs. As shown \Cref{table: self-reflection acc}, we notice that the improvements are not significant, and occasionally detrimental. For these incorrect initial responses, only 6.9\% of cases are detected. 
Our initial experiments ($\mathsection$~\ref{sec:performance_delta}) indicate that intrinsic reflection has limited effect. Across various LLMs and tasks, the performance gains from reflection are not significant, and occasionally detrimental. In cases of incorrect initial responses, only 15.1\% of incorrect responses are corrected through reflection. 
To ascertain the reasons for that, we further analyze the feedback generated by the self-evaluate process. As shown in \Cref{figure1}, LLMs often provide two unexpected feedback: 
\begin{inparaenum}[1)]
    \item \textit{Overconfidence} (46.7\%): Stubbornly insisting that the previous solution is correct.
    \item \textit{Inconsistency} (45.7\%): The feedback is highly inconsistent when self-evaluating the same response multiple times.
\end{inparaenum}
These two feedbacks seriously undermine the effectiveness of reflection. It reveals that such a simple self-evaluate strategy faces difficulty in accurately identifying errors and consistently generating high-quality feedback for reflection.

%
%Those "stubborn" or "random" feedbacks significantly impair reflection performance. 

%This "stubborn" and "random" feedback significantly impairs reflection performance. To a

%Our feedback analysis indicates that a simple self-evaluate strategy is challenging to identify errors accurately and consistently. 

% As a remedy, we propose a contrastive strategy to replace directly evaluating: we contrast the differences among multiple responses and draw inspiration from their disparities for reflection.
% The insight is that generating accurate feedback directly may be difficult, but contrasting the differences between multiple responses seems easier.
As a remedy, we propose a contrastive strategy as an alternative to the direct evaluation: we examine the differences among multiple responses and draw inspiration to derive feedbacks from their disparities for reflection.
The insight is that while generating accurate feedback directly may be challenging, identifying contrasts between various responses is often more feasible.
More importantly, these discrepancies often indicate some potential errors, easily overlooked details or pitfalls. As shown in \Cref{figure1}, by contrasting two solutions, LLM finds they have different solving objectives, and suggests re-examining the intent of the original request in the checklist. This contrasting paradigm can also be seen in some contemporaneous work~\citep{wan2023sequence, yuan2024self}.

%In light of this, we consider that since the role of feedback is to evaluate the correctness of the prior response, could we guide the LLMs to identify potential issues in previous responses by contrasting the discrepancies among multiple solutions? 

% It means LLMs could be inspired by contrasting different solutions.  这种不同solution之间的差异

Embracing this philosophy, we advocate Self-Contrast, which steers LLM to autonomously create diverse solving perspectives by self-curated prompts and then select different results with significant discrepancies for comparison. Then LLM reflects on the reasons behind these discrepancies and generates multiple re-examining instructions, i.e., checklist, for reflection. Our experiments show that by creating diverse perspectives adaptively, Self-Contrast can mitigate biases introduced by specific prompts. Moreover, contrasting the discrepancies between perspectives inspires deeper reflection, thereby enhancing the likelihood of accurate self-correction.

Our contributions can be summarized as:
\begin{itemize}
\setlength{\parskip}{2pt}
\setlength{\itemsep}{0pt plus 1pt}
\item Our comprehensive experiments unveil that the bottleneck for poor reflection performance lies in the LLM's inability to accurately evaluate prior responses. It often manifests as overconfident or inconsistent feedback, hindering the effectiveness of self-reflection.

%We reveal that the quality of feedback is a critical bottleneck for LLM's reflection. However, LLM utilizes inherent ability to consistently provide accurate feedback is challenging, as their feedback is either stubborn or stochastic.

%Reflection tends to be ineffective when LLMs display stubborn behavior due to overconfidence or make stochastic corrections as a result of indecision.

\item We advocate Self-Contrast: the LLMs create multiple solving perspectives for diverse results, mitigating overconfident biases of a singular prompt. Then drawing inspiration from contrasting different perspectives, the LLMs summarize more accurate checking instructions to resolve discrepancies and enhance reflection.

%autonomously contrast the differences among multiple perspectives, providing accurate insight to inspire self-correction.

%LLMs adaptively devise multiple solving perspectives and autonomously contrast their inconsistency for an insightful reflection. 
    
%\item Self-Contrast generates multiple prompts for different perspectives and organizes them into a graph, where each vertex denotes a solution, and the edge signifies the discrepancies between two nodes. Reflection is performed by mitigating the disparities on each edge.

\item Empirically, compared with vanilla reflection, Self-Contrast shows significant improvements and stability in both mathematical reasoning and challenging translation scenarios.

\end{itemize}

\section{Evaluation of Intrinsic Reflection}

%In this chapter, we analyze the intrinsic reflection capabilities of LLMs, including open-source models and commercial APIs, across a diverse array of tasks. 

We first comprehensively investigate the intrinsic reflection capability of LLMs, i.e., LLMs self-evaluate the initial response without external feedback and then refine it. Subsequently, we methodically investigate the factors influencing reflection.

\subsection{Performance Pre- and Post-Reflection}
\label{sec:performance_delta}
%We evaluate reflection  on different tasks using multiple LLMs (GPT4, -3.5, text-davinci-003, Llama2-70b, -13b, -7b). 

We evaluate the reflection capabilities of multiple LLMs across a variety of benchmarks, including math reasoning and creative translation tasks. We report average accuracy for math reasoning and the BLEURT score between predicted sentences and references for the translation task (see $\mathsection$~\ref{experiments setting} for detail). Each result is evaluated multiple times on different prompts. Besides, we also report the significance level (one-tailed t-test) of the accuracy change pre- and post-reflection.

%To mitigate the influence of prompts, we design multiple different prompts for reflection and compute their average performance for pre- and post-reflection. 

As shown in \Cref{table: self-reflection acc}, we observe no significant accuracy changes before and after reflection. For instance, the performance of GPT-3.5 on GSM8K and SVAMP exhibit marginal changes of -0.8\% and +0.7\% after reflection respectively, both statistically insignificant. This negligible performance fluctuation can be validated across multiple LLMs and various benchmarks, far from expectations. Specifically, most reasoning cases suffer from a slight decrease, while the translation task shows little impact. Additionally, smaller LLMs (e.g., Llama2-7B) demonstrate poorer reflection ability, occasionally even exhibiting negative impacts. These experiments collectively suggest that LLMs appear to be incapable of self-correction through reflection.

\begin{table}[t!]\small
\centering
\begin{tabular}{ l | c  c  c}
\toprule[1pt]
 & \multicolumn{2}{c}{\textbf{Math Reasoning}} & \textbf{Translation} \\ 
 & GSM8K & SVAMP & CommonMT\\ \hline

GPT4  & 93.9$\Rightarrow$95.1 & 93.0$\Rightarrow$91.5 & 70.1$\Rightarrow$69.8 \\ 
\color{blue}{P for $\Delta>0$}  & 0.1933 & 0.5846 & 0.5426   \\ \hline

GPT3.5  & 76.6$\Rightarrow$75.8 & 79.8$\Rightarrow$80.5 & 69.1$\Rightarrow$69.3 \\ 
\color{blue}{P for $\Delta>0$}  & 0.6613 & 0.4306 & 0.4420   \\ \hline

davinci-003  & 51.1$\Rightarrow$49.6 & 63$\Rightarrow$63.5 & 62.4$\Rightarrow$63.8 \\ 
\color{blue}{P for $\Delta>0$} & 0.6988 & 0.4729 & 0.2009   \\ \hline

Llama2-70B  & 52.6$\Rightarrow$49.3  & 66$\Rightarrow$63.0 & 63.2$\Rightarrow$62.2\\
\color{blue}{P for $\Delta>0$}  & 0.8416 & 0.9521 & 0.7723 \\ \hline

Llama2-13B  & 28.3$\Rightarrow$29.8  &  42.2$\Rightarrow$42.5  & 62.5$\Rightarrow$61.5\\
\color{blue}{P for $\Delta>0$}  &  0.3855 & 0.2508 & 0.4690   \\ \hline

Llama2-7B & 19.8$\Rightarrow$17.0 & 37.5$\Rightarrow$36.1 & 53.7$\Rightarrow$48.8\\ 
\color{blue}{P for $\Delta>0$}  & 0.9578 & 0.5770 & 0.7492  \\
\bottomrule[0.5pt]
\end{tabular}
\caption{We calculate the average accuracy of the ten experiments for pre- and post-reflection: Pre Acc. $\Rightarrow$ Post Acc. We also report the accuracy change's significance level (P-value) for ten trials, where $\Delta\!=\!Post - Pre$. A larger P indicates a less significant improvement.} \label{table: self-reflection acc}
\end{table}

\subsection{Feedback Analysis}

To investigate the reasons behind the failure of reflection, we further analyze the feedback generated during the self-evaluate process. We classify all samples in GSM8K into four categories based on their correctness of the pre- and post-reflection: 1) \emph{Invalid Reflection} (\color{red}{\ding{55}}\color{black}{$\Rightarrow$}\color{red}{\ding{55}}\color{black}{)} means the results before and after reflection are both incorrect. 2) \emph{Valid Reflection} (\color{red}{\ding{55}}\color{black}{$\Rightarrow$}\color{green}{\ding{51}}\color{black}{)} means a wrong solution is revised to correct through reflection. 3) \emph{Toxic Reflection} (\color{green}{\ding{51}}\color{black}{$\Rightarrow$}\color{red}{\ding{55}}\color{black}{)} represents that an originally correct response is changed to incorrect after reflection. 4) \emph{Others} counts the number of correct $\Rightarrow$ correct. Automatic statistics for the reflection category. 

\textbf{Step 1}: We categorize the reflection into the above four categories. This process can be automated for mathematical benchmarks by comparing whether the answers are correct before and after reflection. For the translation task, we leverage GPT-4 along with annotated answers to evaluate the accuracy of translation results before and after reflection. \textbf{Step 2}: We manually assess the quality of the feedback generated in each reflection case (Invalid, Valid, and Toxic). Based on the correctness and consistency of these feedbacks, we categorize them into four cases (inconsistent, overconfident, etc.). The detailed results are as follows:

\textbf{Fail to Correct the Wrong Initial Response.} As shown in~\Cref{feedback_analyze}, we observe the number of \emph{Toxic Reflection} (\color{green}{\ding{51}}\color{black}{$\Rightarrow$}\color{red}{\ding{55}}\color{black}{: 52)} and \emph{Valid Reflection} (\color{red}{\ding{55}}\color{black}{$\Rightarrow$}\color{green}{\ding{51}}\color{black}{: 48)} are nearly similar. This explains why there is no discernible difference in performance pre- and post-reflection. Besides, considering the scenario when the initial response is erroneous, we observe the number of \emph{Invalid Reflection} (\color{red}{\ding{55}}\color{black}{$\Rightarrow$}\color{red}{\ding{55}}\color{black}{: 269)} is significantly larger than \emph{Valid Reflection} (\color{red}{\ding{55}}\color{black}{$\Rightarrow$}\color{green}{\ding{51}}\color{black}{: 48)}, which indicates LLM fails to correct errors in the initial responses for most cases.

\textbf{Often Provide Overconfident or Inconsistent Feedback.} We examine whether LLMs could generate feedback accurately and consistently. For each sample, we instruct the LLM to evaluate its initial response multiple times and record multiple feedbacks. We manually assess the consistency and correctness of these feedbacks and then summarize each sample into 4 cases: \uppercase\expandafter{\romannumeral1}. \emph{Accurately identifies errors}: In multiple repeated evaluations, the LLM identifies errors and provides accurate and consistent feedback. \uppercase\expandafter{\romannumeral2}. \emph{Stubbornly offers erroneous feedback}: The majority of evaluations provide incorrect feedback with specific errors. \uppercase\expandafter{\romannumeral3}. \emph{Can not output consistent feedback}: Unable to assess consistently, as most feedback is different and quite random for a same initial response. \uppercase\expandafter{\romannumeral5}. \emph{Overconfidence, no revision required}: LLM is overconfident and believes no revision is necessary. The detailed evaluation criteria are provided in~\Cref{Detail For Manual evaluation}.

%Subsequently, we manually evaluate the consistency and correctness of these feedbacks for each sample (369 samples in total). 

As shown in \Cref{feedback_analyze}, for the majority of Invalid  Reflection, their feedback is either overconfident (53.5\%) or highly inconsistent (45.3\%), making it difficult to prompt reliable reflection. Similarly, in Toxic Reflection scenarios, 65.4\% of the evaluation processes are highly inconsistent, leading to many correct answers being erroneously modified.

%for most reflection cases, the feedback generated by LLM itself is either highly inconsistent and random (Item 3: about 40\%), or it exhibits overconfidence in the previous response (Item 4, 5: about 50\%). Both cases are highly conducive to the failure of the reflection.

\begin{table}[t!]\small
    \centering
    \begin{tabular}{c|lccc} \toprule[1pt]
         & \multirow{2}*{\makecell[l]{\#Invalid: 269\\ \#Valid: 48\\ \#Toxic: 52}}&  \multicolumn{3}{c}{\textbf{Reflection Behavior}}\\ \cline{3-5}  
         &  &  \makecell[c]{Invalid\\ \scriptsize\color{red}{\ding{55}}\color{black}{$\Rightarrow$}\color{red}{\ding{55}}} &  \makecell[c]{Valid\\ \scriptsize\color{red}{\ding{55}}\color{black}{$\Rightarrow$}\color{green}{\ding{51}}}& \makecell[c]{Toxic\\\scriptsize\color{green}{\ding{51}}\color{black}{$\Rightarrow$}\color{red}{\ding{55}}}\\ \hline 
        \textbf{\multirow{4}*{\rotatebox{270}{Feedback Type}}} &  \makecell[l]{I. Accurately\\identifies errors}&0.4\% & 43.3\% & 0\%\\ \cline{2-2}
         &  \makecell[l]{II. Stubbornly offers\\erroneous feedback}& 0.8\% & 0\% &31.1\% \\ \cline{2-2} 
         &  \makecell[l]{III. Can not output \\consistent feedback}& 45.3\% & \textbf{47.5\%} & \textbf{65.4\%}\\ \cline{2-2} 
         &  \makecell[l]{IV. Overconfidence\\No revision required}& \textbf{53.5\%} & 9.2\% & 3.5\%\\ \bottomrule[1pt]
    \end{tabular}
    \caption{We consider three reflection behaviors based on the correctness of the pre- and post-reflection: Invalid, Valid, and Toxic. Besides, we summarize each sample's feedback into four categories when self-evaluation.}\label{feedback_analyze}
\end{table}

\subsection{From Self-Evaluate to Self-Contrast}\label{section2.3}

%The experiments above suggest that LLM struggles to self-evaluate prior solutions accurately and may generate some overconfidence or inconsistent feedback. 
The aforementioned experiments indicate that feedback generated by the self-evaluate process is either highly random or excessively confident. This unstable self-evaluate may severely impact the reflection performance of LLMs.

%This suggests that LLMs exhibit cognitive biases or significant uncertainties in their self-assessment of certain samples, 

%making it challenging to prompt effective self-reflection in these scenarios.

As a remedy, we propose a contrastive strategy for reflection. Instead of directly evaluating a response, which can be challenging and inconsistent, we instruct the LLM to initially contrast the differences between various solutions, and identify their discrepancies and the reasons behind them. As shown in \Cref{figure1} (bottom), we sample Top-2 responses from LLM and then prompt LLM to contrast their differences in detail, rethink the reasons that caused the discrepancies, and summarize the checklist for re-examining and resolving the discrepancy. 
As shown in \Cref{indirect feedback}, we compare three scenarios: self-evaluate w/ top-1 response, self-evaluate w/ top-2 responses, and self-contrast w/ top-2. Our new strategy achieves a modest improvement over standard reflection using self-evaluate. Notably, it significantly enhances the significance levels (p-value: 0.6613 to 0.0933), suggesting it can greatly mitigate the self-evaluation process's uncertainty.

 In this section, we validate the concept of contrastive evaluation. For the next section, we expand this contrastive concept into full-version self-contrast, which involves creating multiple perspectives, and contrasting their differences, summarizing the checklist for deeper reflection.
\begin{figure*}[!t] %H为当前位置，!htb为忽略美学标准，htbp为浮动图形
\centering %图片居中
\includegraphics[width=1\textwidth]{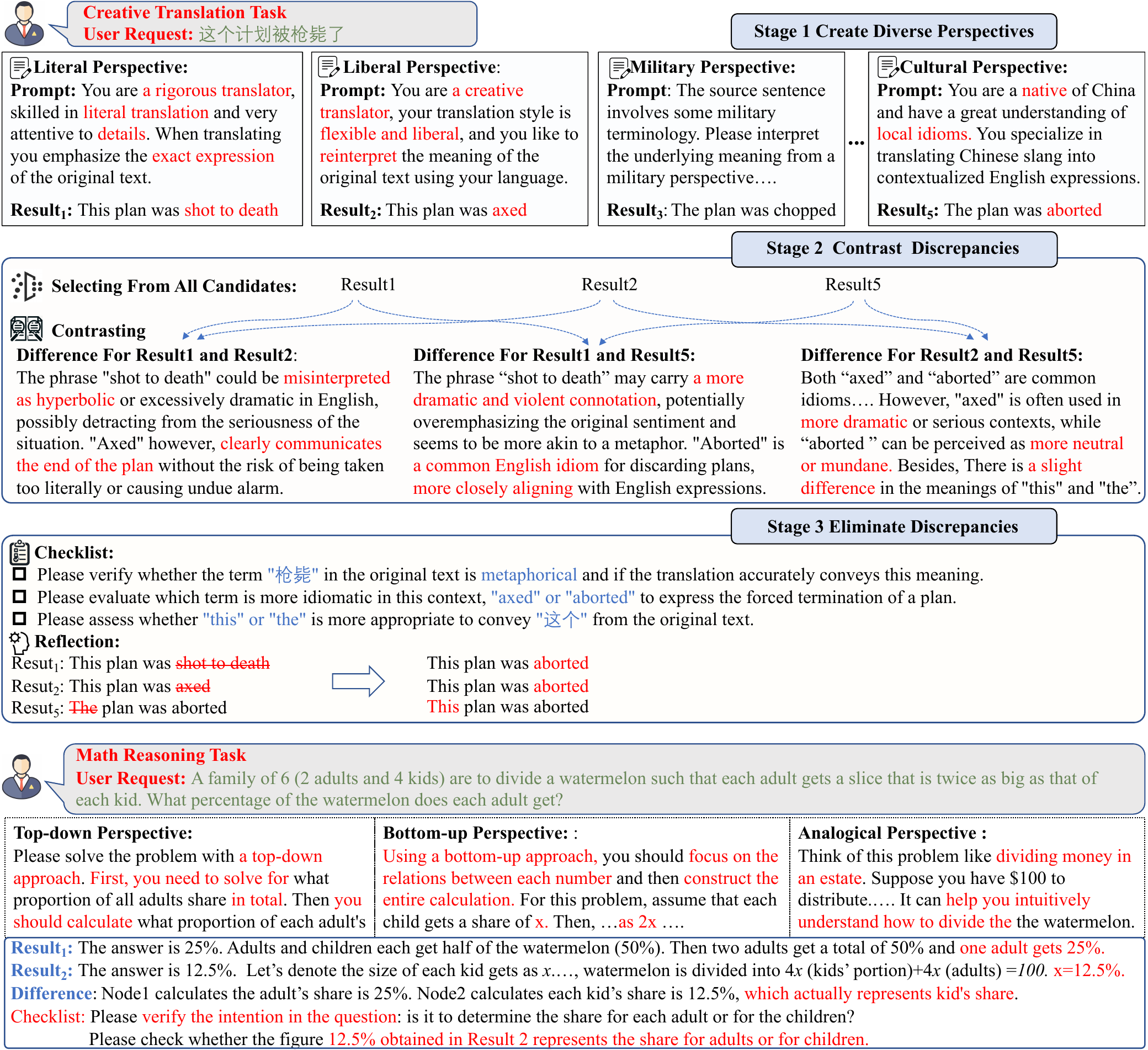} %
\caption{Self-Contrast designs diverse prompts for different solving perspectives and generates corresponding results. Then we filter out similar results and select those that are significantly different. To inspire reflection, we contrast the differences between selected results and prompt LLM to summarize a checklist. This checklist can be used to re-examine and eliminate discrepancies. Lastly, LLM revises each response to achieve a consistent result.}\label{figure2} %最终文档中希望显示的图片标题
\end{figure*}
%Self-Contrast首先根据user request自适应地设计多个不同视角的prompts (Step1), 然后生成多个候选结果并当作nodes(Step2). 此外一个Intentionally Incorrect Perspective生成的结果被当作anchor. 我们通过K-Medoids算法将多个候选节点聚类成三个类别(Step3)。然后基于他们和锚点的相似度，保留每个类别中最不相似的节点(Step4)。此后Self-Contrast对比节点间的差异并构建关系图 (Step5)。最后LLM不断修正node来消除差异，从而实现反思(Step6).Moreover, a result generated from \emph{Incorrect Perspective} serves as an anchor

\begin{table}[t!]\small
\centering
\setlength\tabcolsep{3pt} % 默认值通常是6pt
\begin{tabular}{ l  c  c  c}
\toprule[1pt]
\textbf{\makecell[l]{Strategy}} & \textbf{\makecell[c]{GSM8K}} & \textbf{\makecell[c]{SVAMP}}& \textbf{\makecell[c]{CommonMT}}\\ \hline
Self-Evaluate w/ top-1 & -0.8 & 0.7 & 0.2\\ 
P for $\Delta\!>\!0$  & 0.6613 & 0.4306 & 0.4420\\ \hline

Self-Evaluate w/ top-2 & 0.12 & 0.8 & 0.16\\
P for $\Delta\!>\!0$  & 0.4192 & 0.3457 & 0.3745\\ \hline

Self-Contrast w/ top-2 & 0.9 & 2.5 & 0.45\\
P for $\Delta\!>\!0$  & \textbf{0.0933} & \textbf{0.0118} & \textbf{0.0457}\\ 
\bottomrule[0.5pt]
\end{tabular}
\caption{We report the accuracy change ($\Delta$) between post- and pre-reflection for 3 settings and t-test value for $\Delta$>0. Self-evaluate: Directly evaluate the initial response. Self-contrast: Contrast the difference between two responses and generate a checklist for reflection.}\label{indirect feedback}
\end{table}

\section{Self-Contrast}
Prior sections illustrate the challenges LLMs encounter in accurately evaluating previous solutions, often resulting in overconfident or inconsistent feedback. Concurrently, we observe that leveraging the discrepancies between two different solutions can inspire a more efficacious reflection, notably reducing the uncertainty during the reflection. Building upon this insight, we propose a more diverse inter-perspective Self-Contrast, facilitating more reliable self-reflection.

%we design a flexible contrastive strategy for self-correction, leveraging discrepancy across multiple problem-solving perspectives.

Self-Contrast consists of three procedures: Create Diverse Perspectives, Contrast Inter-Perspective Discrepancies, and Eliminate Discrepancies. In Create Diverse Perspectives ($\mathsection$~\ref{section 3.1}), we encourage LLMs to autonomously create a variety of prompts tailored to the user's request, each offering a unique perspective for problem-solving, e.g., different thinking styles, diverse identities, personalities, or preferences. These diverse perspectives prompt the LLM to generate different responses. In the second stage ($\mathsection$~\ref{section 3.2}), LLM contrasts the differences between each pair of responses. Lastly ($\mathsection$~\ref{section 3.3}), to eliminate discrepancies, we abstract these differences into a detailed checklist for re-examining. This checklist guides the LLM to meticulously examine the causes of discrepancies, including random errors or intrinsic biases, which result in inconsistent results among perspectives.

As shown in \Cref{figure2}, LLM designs five different prompts and their translation results based on the user's request \begin{CJK*}{UTF8}{gkai}("这个计划被枪毙")
\end{CJK*}. From a literal perspective, the phrase \begin{CJK*}{UTF8}{gkai}"被枪毙"\end{CJK*} is translated as "shot to death". This rigid translation fails to grasp the metaphor embedded in the military term. Conversely, from a liberal perspective, it is translated as "This plan was axed". After contrasting two different translations, LLMs believe they should scrutinize the source sentence for metaphors and ensure the translation aligns with the conventions of English expression.

%each focusing on a unique mindset and emphasis. Then, LLM

%LLM meticulously examines the lexical and semantic differences between $result_1$ and $result_5$, concluding that $result_1$ is inappropriate. For math reasoning, the LLM discovers that different answers are caused by different understandings of the problem's intent. 

%and then organize them into a graph, where each node denotes a result, and each edge documents the differences and checklists for connected nodes. 

%The reflection process is reformulated as graph editing, i.e., continuously editing inconsistent nodes to eliminate discrepancies.

%Then we employ clustering algorithms to select the results that exhibit significant discrepancies from all perspectives. 

%each perspective with customized roles, personalities, and mindsets
 
\subsection{Create Diverse Perspectives} \label{section 3.1}
\textbf{Self-Curated Prompts}
First, it is imperative to define the concept of "solving perspective". It refers to deliberate prompting with a unique role, personality, thought style, etc., which prompts LLMs to solve user requests from a specific perspective. Diverse solving perspectives can endow LLMs with a broader range of thoughts for problem-solving, e.g., different angles and methodologies, thereby mitigating biases introduced by singular prompts. 

% To achieve this, we adopt a self-curated prompt strategy to adaptively generate multiple different prompts for each request by LLM itself, each signifying a tailored perspective, then sample corresponding responses based on these prompts.
To achieve this, we adopt a self-curated prompt strategy, where the LLM itself adaptively generates multiple different prompts for each request, each signifying a tailored perspective, then samples corresponding responses based on these prompts.
% It is noteworthy that how many perspectives should be created, and how to design each perspective, are entirely determined by LLMs, which endows LLMs with more flexibility to address complex tasks.
It is noteworthy that the number of perspectives to be created, and the design of each perspective are entirely determined by LLMs, endowing them with more flexibility to address complex tasks.
The details of the prompt are provided in \Cref{prompt for self-curated}. In \Cref{self-curated results}, we present statistics on the number of prompts generated in self-curated prompt process.

%In this context, we categorize existing prompting methods from two dimensions: \emph{Fixed or dynamic perspectives? Singular or multiple perspectives?} As shown in~\Cref{prompting strategy}, most existing strategies adopt singular or fixed prompts. 

%However, our strategy adopts a self-curated strategy to create multiple dynamic perspectives without the need for manually crafted prompts, 

%a singular prompt often struggles with complex problems and fixed prompts require manual crafting. 

\iffalse
\textbf{Intentionally Incorrect Perspective} 
In addition to the above self-curated perspectives, we also introduce an imperfect or even incorrect solving perspective. Specifically, we instruct LLMs to consider what common errors might be made for this request, then actively adopt a careless persona to generate an incorrect response with some common mistakes. It can serve as a negative demonstration for subsequent selection and reflection.

\fi

\subsection{Contrast Inter-Perspective Discrepancies} \label{section 3.2}
%We employ a relational graph to depict key responses and the differences between them. 
The LLM generates diverse responses based on self-curated prompts, each representing a specific perspective. Considering that some responses may be highly similar or even identical, we first filter these similar responses. Then, we select the responses with significant discrepancies for comparison.

%To effectively contrast discrepancies across perspectives,

%Owing to the token number constraints within LLMs, constructing a fully connected graph for all nodes is impractical. 

%Therefore, we first aggregate all nodes into multiple clusters based on their semantic similarity and then select the nodes with obvious differences to construct the graph.

\textbf{Selecting} To filter out similar responses, we employ the K-Medoids clustering algorithm based on their semantic similarity. We categorize all responses into $k$ clusters, each encompassing a set of similar results. Then we select the centroids of each cluster as representative responses and discard the remaining ones. It ensures the selected results exhibit substantial differences from each other.

%, while simultaneously having a higher probability of being correct compared to other responses.

%Dissimilar responses are separated from each other.

%To select responses that exhibit significant discrepancies, we select only one response from each category. 

%Specifically, we first treat the result of \emph{Intentionally Incorrect Perspective} (delineated in~\Cref{section 3.1}) as an anchor. This response contains some common mistakes created by LLM itself. So it can serve as a counterexample to indicate some potential errors. We compute the semantic similarity between each response and the anchor. 

%Then for each cluster, we select the response that exhibits the least similarity to the anchor (Step4 in \Cref{figure2}). 

\textbf{Contrasting} After selecting $k$ responses from all candidates, we feed these responses concurrently into LLM and then instruct LLM to autonomously contrast the differences for each pair of responses in a single pass. When contrasting, LLMs need to explicitly answer these questions: \textbf{Whether} the two responses are different, \textbf{Where} the differences lie, and \textbf{Which factors} contributed to these inconsistent results. These questions guide the LLM to methodically explore the underlying reasons behind discrepancies, identifying potential errors and often overlooked details. As shown in \Cref{figure2}, for translation tasks, the LLM compares results 1, 2, and 5, and identifies that their primary differences lie in the use of different verbs to express \begin{CJK*}{UTF8}{gkai}"被枪毙"\end{CJK*}. The detailed prompts are shown in \Cref{edge generating}. 

%we obtain $\tbinom{k}{2}$ differences.
%Specifically, we feed $k$ responses concurrently into LLM and then instruct LLM to contrast the differences for each pair of responses ($\tbinom{k}{2}$ in total). 

\subsection{Eliminate Discrepancies} \label{section 3.3}
We abstract insightful checklists from these pairwise contrastive differences and then use them to resolve the inconsistencies across various perspectives for a consensus.

%Initially, we instruct the LLM to summarize checklists from these identified differences, followed by directing the LLM to amend the corresponding results from a global perspective, thereby achieving consensus.

\textbf{Summarizing Checklist} 
To ascertain the truth and resolve discrepancies, the LLM is encouraged to summarize a detailed checklist for re-examining the user's request and candidate responses. This checklist contains multiple specialized checking instructions, such as verifying alignment with the user's intent, identifying contradictions in LLM's responses, checking for miscalculations, etc. It explicitly points out some potential issues, e.g., previously overlooked details, logical pitfalls, or unreasonable steps, and compels LLM to re-examine them. Compared to conventional reflection instruction, e.g., \emph{Please check your previous response}, our checklist is more precise and informative.

%re-examine the user's request and the previous results, avoiding pitfalls and common errors. 

\textbf{Reflection For Consensus}
Lastly, we employ the checklist and identified discrepancies to prompt reflection. LLM can revise the inconsistent perspectives and output $k$ consistent responses.

%Firstly, we employ a JSON format in prompt for $k$ candidates and $\tbinom{k}{2}$ differences, i.e., each result acts as a node and each edge denotes the differences between two connected nodes: \emph{Node: \{\{result1\}, \{result2\}, \{result3\}..\}, Edge: \{\{difference1\}, \{difference2\}..\}}. It provides a more intuitive and global view for reflection. Besides, we also provide the generated checklist and an imperfect result (from the Incorrect Perspective) in the prompt to guide reflection. LLM is prompted to revise inconsistent steps of all nodes, eliminating their discrepancies, and outputting a new graph with all consistent nodes. To reach a consensus among perspectives, LLM requires careful and comprehensive consideration, as even minor modifications may lead to new discrepancies with others. We provide a complete prompt in \Cref{prompt format}. 
Concretely, we use a JSON format for the revision prompt: \emph{Request: \{\{request\}\}, Candidate: \{\{result1\}, \{result2\}, \{result3\}..\}, Discrepancy: \{\{difference1-2\}, \{difference1-3\}..\}, Checklist: \{\{instruction1\},\{instruction2\},..\}}. To eliminate discrepancies, we instruct LLM to revise the inconsistent steps of each candidate and output $k$ revised responses with consistent answers. When revising, LLM should require careful and comprehensive consideration, as any minor modifications may lead to new discrepancies with others.

%It provides a more intuitive and global view for reflection. 

%Notably, to eliminate inconsistencies on all edges, LLMs must reflect globally and cautiously edit problematic nodes, resolving the conflicts between each two nodes. Since minor edits to any node may potentially give rise to new inconsistent edges.

%Lastly, the editing process stops when the LLM successfully revises all problematic nodes, achieving a harmonized graph represented by $k$ consistent nodes. If any inconsistencies persist, the above process is repeated until convergence is achieved.

%Each edge furnishes key differences between two candidate results from different perspectives. 

%These differences can be viewed as a form of external feedback, despite being internally generated by LLMs themselves. 

%This self-generated "external" feedback can elicit LLMs to overcome their own stubborn beliefs and inspire LLMs to rethink which solution is correct.

\section{Experiments}

\begin{table*}[t!]\small
\centering
\setlength\tabcolsep{3pt} % 默认值通常是6pt
\begin{tabular}{l|lllll|lllll|c}
\toprule[1pt]
& \multicolumn{5}{c|}{GSM8K} & \multicolumn{5}{c|}{SVAMP} &\multirow{2}*{\makecell[c]{\#Call\\Avg.}}\\
\cmidrule(lr){2-6} \cmidrule(lr){7-11}
& GPT3.5 & GPT4 & L-7B & L-13B & L-70B  & GPT3.5 & GPT4 & L-7b & L-13B & L-70B &  \\
\midrule 
CoT Prompt  &  76.6   &  93.9  & 19.8   & 28.3  & 52.6 &  79.8    & 93.0   & 37.5   & 40.2  & 66 & 1\\ \hline 
ExpertPrompt     &  77.3 \tiny{$\uparrow$0.7}   &  93.8 \tiny{$\downarrow$0.1}  & 21.6 \tiny{$\uparrow$1.8}   & 
30.5 \tiny{$\uparrow$2.2}  & 53.1 \tiny{$\uparrow$0.5} &  80.2 \tiny{$\uparrow$0.4}    &  93.3 \tiny{$\uparrow$0.3}  & 37.7 \tiny{$\uparrow$0.2}   &  41.9 \tiny{$\uparrow$1.7}  &65.6 \tiny{$\uparrow$0.4} & 2\\  
Self-Reflection   & 75.8 \tiny{$\downarrow$0.8} & 95.1 \tiny{$\uparrow$1.2}   & 17.0 \tiny{$\downarrow$2.8}   &  31.8 \tiny{$\uparrow$3.5}  & 49.3 \tiny{$\downarrow$3.3}  & 80.5 \tiny{$\uparrow$0.7} &  91.5 \tiny{$\downarrow$1.5}    & 36.1 \tiny{$\downarrow$1.4}   &  42.5 \tiny{$\uparrow$2.3}  &  63.0 \tiny{$\downarrow$3}  &  3\\  
Self-Consistency    &     &   &   &  &  &    &   &   &   &  & \\

-- SC-Vote    &  83.5\tiny{ $\uparrow$6.9}    & 94.2\tiny{ $\uparrow$0.3}   &  21.4\tiny{ $\uparrow$1.6}  & 37.6\tiny{ $\uparrow$9.3}  & 61.1\tiny{ $\uparrow$8.5} &  84.6\tiny{ $\uparrow$4.8}   &  92.5\tiny{ $\downarrow$0.5}  & \textbf{45.2}\tiny{ $\uparrow$\textbf{7.7}}   &  53.7\tiny{ $\uparrow$13.5}  & 72\tiny{ $\uparrow$6} & 8\\  

-- SC-Select    &  76.3\tiny{ $\downarrow$0.3}   &  93.1\tiny{ $\downarrow$0.8}  & 16.2\tiny{ $\downarrow$3.6}   & 28.6\tiny{ $\uparrow$0.3}  & 54.6\tiny{ $\uparrow$2.0} &  81.2\tiny{ $\uparrow$1.4}    & 93.2\tiny{ $\uparrow$0.2}   & 35.1\tiny{ $\downarrow$2.4}   & 38.9\tiny{ $\downarrow$1.3}  & 66.5\tiny{ $\uparrow$0.5} & 9\\

-- SC-Reflect     &  75.8 \tiny{$\downarrow$0.8}  &  93.3 \tiny{$\downarrow$0.6} & 19.2 \tiny{$\downarrow$0.6}  & 29.1 \tiny{$\downarrow$0.8}  & 53.7 \tiny{$\uparrow$1.1} &  81.1 \tiny{$\uparrow$1.3}   & 93.4 \tiny{$\uparrow$0.4}  & 32.5 \tiny{$\downarrow$5}   & 34.2 \tiny{$\downarrow$6}  & 67.5 \tiny{$\uparrow$1.5} & 9\\ 

Multi-Agent  &  83.8 \tiny{$\uparrow$7.2}   & 93.5 \tiny{$\downarrow$0.4}  &  \textbf{23.8} \tiny{$\uparrow$\textbf{4}} & 34.9 \tiny{$\uparrow$6.6}  & 59.6 \tiny{$\uparrow$7.0} &  84.1 \tiny{$\uparrow$4.3}  &  93.2 \tiny{$\uparrow$0.2} & 42.5 \tiny{$\uparrow$5}  &  49.2 \tiny{$\uparrow$9.0}  & 70.1 \tiny{$\uparrow$4.1} & 9\\  \hline 

Hint-Prompt     & 78.8 \tiny{$\uparrow$2.2}  &  93.7 \tiny{$\downarrow$0.2} &  18.3 \tiny{$\downarrow$1.5} & 27.8 \tiny{$\downarrow$0.5}  & 59.6 \tiny{$\uparrow$7} &   79.3 \tiny{$\downarrow$0.5}  &  93.1 \tiny{$\uparrow$0.1} & 38.8 \tiny{$\uparrow$1.3}  &  40.6 \tiny{$\uparrow$0.4} &67.6 \tiny{$\uparrow$1.6} & 6.7\\  

Math-Prompt    & 79.6 \tiny{$\uparrow$3.0}   &  93.9 \tiny{$\downarrow$0.0} & 19.5 \tiny{$\downarrow$0.3}  & 30.6 \tiny{$\uparrow$2.3}  & 59.8 \tiny{$\uparrow$7.2}  &     81.2 \tiny{$\uparrow$1.4} &  93.6 \tiny{$\uparrow$0.6} &  37.2 \tiny{$\downarrow$0.3} & 41.5 \tiny{$\uparrow$1.3}   & 68.7 \tiny{$\uparrow$0.5} & 4.5 \\ \hline

Self-Contrast  &  \textbf{84.4} \tiny{$\uparrow$\textbf{7.8}}   &  \textbf{95.4} \tiny{$\uparrow$\textbf{1.5}}  & 20.5 \tiny{$\uparrow$0.7}   & \textbf{42.3 }\tiny{$\uparrow$\textbf{9.2}}  & \textbf{64.2} \tiny{$\uparrow$\textbf{11.6}} &   \textbf{89.0} \tiny{$\uparrow$\textbf{9.2}}   &  \textbf{94.0} \tiny{$\uparrow$\textbf{1}}  &  44.5 \tiny{$\uparrow$7}  &  \textbf{54.6} \tiny{$\uparrow$\textbf{14.4}}  & \textbf{75.3} \tiny{$\uparrow$\textbf{9.3}} & 7.8 \\

\bottomrule[0.5pt]
\end{tabular}
\caption{The performance on mathematical reasoning. Self-Consistency (SC-Vote, -Select, -Reflect) samples eight responses and then performs voting, selecting, or reflection. For the Multi-Agent, we configure three agents to engage in a three-round debate. $\uparrow$ and $\downarrow$ means accuracy changes over the CoT prompt. \texttt{L-} denotes Llama2-chat.} \label{combined_main_result}
\end{table*}

\begin{table}[t!]\small
\centering
\setlength\tabcolsep{4.2pt} % 默认值通常是6pt
\begin{tabular}{l|cccc}
\toprule[1pt]
&\makecell[c]{GPT3.5}  & \makecell[c]{L-7B} & \makecell[c]{L-13B} & \makecell[c]{L-70B} \\
\hline  
CoT Prompt    &  69.1   &    53.7    &  62.5  &  63.2  \\\hline
ExpertPrompt          & 69.6 \tiny{$\uparrow$0.5} & 53.8 \tiny{$\uparrow$0.1} & 62.9 \tiny{$\uparrow$0.4} & 63.4 \tiny{$\uparrow$0.2} \\    
Self-Reflection          & 69.3 \tiny{$\uparrow$0.2} & 48.8 \tiny{$\downarrow$4.9} & 61.5 \tiny{$\downarrow$1.0} & 62.2 \tiny{$\downarrow$1.0} \\  
Self-Consistency &&&&\\
-- SC-Vote      & -- & -- & -- & -- \\  
-- SC-Select             & 68.6 \tiny{$\downarrow$0.5} & 52.1 \tiny{$\downarrow$1.6} & 62.8 \tiny{$\uparrow$0.3} & 63.0 \tiny{$\downarrow$0.2} \\  
-- SC-Reflect            & 69.0 \tiny{$\downarrow$0.1} & 54.0 \tiny{$\uparrow$0.3} & 62.2 \tiny{$\downarrow$0.3} & 63.2 \tiny{$\uparrow$0} \\  
Multi-Agent           & 69.9 \tiny{$\uparrow$0.8} & 51.9 \tiny{$\downarrow$1.8} & \textbf{63.1} \tiny{$\uparrow$\textbf{0.6}} & 65.8 \tiny{$\uparrow$2.6} \\  
Hint-Prompt           & 69.6 \tiny{$\uparrow$0.5} & \textbf{54.2} \tiny{$\uparrow$\textbf{0.5}} & 62.5 \tiny{$\uparrow$0} & 64.6 \tiny{$\uparrow$1.4} \\ \hline
Self-Contrast         & \textbf{70.7} \tiny{$\uparrow$\textbf{1.6}} & 52.1 \tiny{$\downarrow$1.6} & 62.8 \tiny{$\uparrow$0.3} & \textbf{66.7} \tiny{$\uparrow$\textbf{3.5}} \\ 
\bottomrule[0.5pt]
\end{tabular}
\caption{The performance on Creative Translation.} \label{commonMT_result}
\end{table}
%For non-numerical tasks, the Self-Consistency is unable to yield a final answer through voting. So we replace with two variant strategies, i.e. SC-Select and SC-Reflect

\subsection{Settings} \label{experiments setting}
\textbf{Benchmarks} We evaluate our method within two testbeds: mathematical reasoning and translation using GSM8K, SVAMP, and CommonMT benchmarks. Please see \Cref{benchmarks} for details.
%We also conduct a comprehensive investigation of the advanced capabilities and limitations inherent in Self-Contrast.

\textbf{Evaluation Metrics} For mathematical reasoning, we evaluate the precision of the final answer after their step-by-step reasoning, similar to the previous methodologies. For the translation task, we employ BLEURT\footnote{\url{https://github.com/google-research/bleurt, BLEURT-20}} score as automatic metrics. 
%In addition, we adopt GPT-4 as an evaluator to compare which of the two models' responses more closely aligns with the annotated translations.

\textbf{LLM Models and Prompts} We conduct experiments using the GPT-3.5-Turbo-0613 and GPT-4-0613, alongside the Llama2-Chat model with three parameter scales (7B, 13B, and 70B). To make a fair comparison, we uniformly set the temperature to 0.2 for all experiments. For standard prompts and self-reflection baseline, we evaluate them 10 times using different prompts and average their results under zero-shot scenes. Prompts and other details can be found in \Cref{baseline_prompt,our prompts,other details}.

%In addition, we also employ GPT-4 to determine who yields superior translation results between the two models.

\subsection{Baselines}
We compare Self-Contrast with the following baselines: Standard CoT Prompt~\citep{kojima2022large}. Self-Reflection~\citep{Shinn2023ReflexionAA}. Multi-Agent Debate~\citep{Du2023ImprovingFA, Liang2023EncouragingDT, he-etal-2020-box}. ExpertPrompt~\citep{Xu2023ExpertPromptingIL}. Hint-Prompt~\citep{Zheng2023ProgressiveHintPI}. Math-Prompt~\citep{Imani2023MathPrompterMR}. Moreover, for various task scenarios, we consider three forms of Self-Consistency~\citep{Wang2023Self, Chen2023UniversalSF}: \textbf{SC-Vote}: The original Self-Consistency version, which samples $K$ decoding results, followed by a voting process. \textbf{SC-Select}: Instead of voting, LLM also samples Top-K responses but then selects the most appropriate answer from $K$ candidates by itself. \textbf{SC-Reflect}: After sampling Top-K responses, LLM reflects on these candidates and regenerates a new response as the final answer.
%~\citep{Yoran2023AnsweringQB}
%Additionally, we combine the idea of self-consistency (-SC) with post-hoc correction, and design two baselines:

\subsection{Main Results}
%三张表格和一个胜率对比图
In \Cref{combined_main_result,commonMT_result}, we report the accuracy and the average number of API/LLM calls (\#Call), which serves as a proxy for the computational cost. 

\textbf{Consistent improvement over vanilla reflection.} Compared to vanilla reflection, Self-Contrast brings significant and stable improvement. For mathematical reasoning, we achieve an average improvement of +7.2\%. In contrast, the original self-reflection shows no clear improvement (-0.51\%). A similar phenomenon is observed in creative translation, where Self-Contrast achieves a +0.95 improvement, whereas self-reflection results in a decrease of -1.6. Besides, compared to multi-agent and ensemble baselines, our improvement is also pronounced and consistent.

\textbf{Better generality across different LLMs and tasks.} From commercial LLMs (e.g., GPT4) to open-source models (Llama-2), and from reasoning to generative tasks, our strategy exhibits robust generalizability. Concretely, from the perspective of LLM, Self-Contrast achieves the best results on most models except Llama-2-7B. For instance, for GPT-3.5, the improvements are 7.8\% on GSM8K and 9.2\% on SVAMP, while for Llama-2-70B, the improvements are 11.6\% and 9.3\% respectively. As for Llama-2-7B, our performance is slightly lower than Self-consistency and Multi-Agent. This might be due to the weaker instruction-following capabilities of the Llama2-7B, making it challenging to contrast two inconsistent solutions. Besides task-wise, Self-Contrast applies to various task types, demonstrating high versatility. In contrast, Self-Consistency can not handle non-numerical tasks directly, e.g., translation, due to its voting mechanism (\Cref{commonMT_result}). Its variant strategies, SC-Select and SC-Reflect, lag significantly behind ours.

%These results suggest that contrasting multiple inconsistent perspectives can mitigate overconfidence and enhance the certainty of reflection.

\textbf{Fewer manual efforts and more reasonable call overheads.} Compared to the multi-agent debate, Self-Contrast gains more significant improvements with less call overhead (>10\% reduction). From a unified perspective, it can be viewed as a multi-agent contrastive mechanism. Instead of a free-form debate among multiple agents, our strategy fosters a more explicit and purposeful debate by contrasting the differences between agents and summarizing the reasons for their disagreements. Moreover, Self-Contrast is flexible, dynamically designing multiple perspectives tailored to user requests, without the need for manually pre-configuring agent roles and quantities.

%It involves grouping and contrasting multiple agents based on their relations and then constructing an agent relation graph. 
%Our explicit contrasting strategy serves as a specific freeform debate among agents.

\iffalse
\begin{table}[t!]\small
\centering
\begin{tabular}{ l | c | c | c}
\toprule[1pt]
\makecell[l]{Variant Strategy}  & \makecell[c]{GPT3.5} & \makecell[c]{Llama2-70B}& \makecell[c]{Llama2-13B}\\ \hline
\makecell[l]{\emph{CoT Prompt} } & 76.6 & 53.1 & 28.3\\ 
\makecell[l]{\emph{Self-Contrast} } & 84.4 & 74.2 & 42.3\\ 
\makecell[l]{\emph{w/o Contrast Edge} } & 78.2 & 59.6 & 29.9\\ 
\makecell[l]{\emph{w/o Graph Form} }& 82.9 & 63.3 & 33.5\\   
\makecell[l]{\emph{w/o Incorrect View} } & 83.6 & 70.3 & 39.8\\ 
\bottomrule[0.5pt]
\end{tabular}
\caption{GSM8K. \label{ablation study} }
\end{table}

\begin{table}[t!]\small
\centering
\begin{tabular}{ l | c | c | c}
\toprule[1pt]
\makecell[l]{Variant Strategy}  & \makecell[c]{GPT3.5} & \makecell[c]{Llama2-70B}& \makecell[c]{Llama2-13B}\\ \hline
\makecell[l]{\emph{CoT Prompt} } & 69.1 & 63.2 & 62.5\\ 
\makecell[l]{\emph{Self-Contrast} } & 70.7 & 66.7 & 62.8\\ 
\makecell[l]{\emph{w/o Contrast Edge} } &  69.5 & 63.3 & 61.8\\ 
\makecell[l]{\emph{w/o Graph Form} }& 70.5 & 65.7 & 62.6\\ 
\makecell[l]{\emph{w/o Incorrect View} } & 69.3 & 64.1 &62.2\\ 
\bottomrule[0.5pt]
\end{tabular}
\caption{CommonMT. \label{ablation study} }
\end{table}

\fi

\section{The Effect of the Different Components}
The above results show that Self-Contrast inspires reflection more accurately and stably than direct evaluation. It encompasses a self-curated prompt process, which fosters diverse solving perspectives to mitigate self-evaluation biases. Besides, it involves a checklist generation process to facilitate re-examination. We analyze their effect as follows:

\textbf{Self-curated Prompt Vs. Sampling Multiple Responses.} Instead of self-curated prompt process, we directly sample multiple responses from LLMs for subsequent contrast and reflection. \Cref{Fig.sample_top-n} shows that the final accuracy improves as the number of sampled responses increases, yet it is still lower than Self-Contrast with self-curated prompts process, where full strategy achieves 84.4\% compared to the maximum of 81.8\% when sampling 5 responses. We find that the top-n responses are sometimes strikingly similar, diminishing the effectiveness of the contrastive strategy.  

\textbf{Reflection Without Checklist.} We eliminate the checklist generation process, i.e., directly instruct the LLM to reflect on the differences among perspectives. In \Cref{reflection on differences}, it brings a significant impact on mathematical reasoning (-3.5\%), but a slight impact on translation (-0.1\%), since translation tasks tend to focus more on local features. Even without a checklist, the LLM also can reflect based on the comparisons of lexical, syntactic.

%We suspect that this is because reasoning tasks are inherently more complex, as abstracting the differences into a checklist can prompt the LLM to rethink from a global perspective.

%\textbf{The Effect of Negative Perspective.} Lastly, we analyze the effect of the "negative" perspective (Intentionally Incorrect Perspective). If we exclude the "negative" perspective and its results from Self-Contrast (we select the centroids of each cluster for \Cref{section 3.2}), we observe a significant decrease in the performance, particularly in translation tasks (in \Cref{reflection on differences}). This suggests that by contrasting positive and negative aspects, this self-generated imperfect result can also inspire deeper reflection.

%as contrast is not only conducted on multiple perspectives but also involves the comparison between "positive" and "negative" perspectives, which also stimulates reflective thinking.

%

%It acts as an erroneous example, guiding the LLM to actively exclude certain common mistakes.

\section{Analysis}
\begin{table}[t!]\small
\centering
\begin{tabular}{ c| l  l  l  }
\toprule[1pt]
\setlength\tabcolsep{1pt} % 默认值通常是6pt
LLMs& Strategy & \makecell[l]{Invalid \color{red}{\ding{55}}\color{black}{$\Rightarrow$}\color{red}{\ding{55}}\\ Cases $\downarrow$} & \makecell[l]{Toxic \color{green}{\ding{51}}\color{black}{$\Rightarrow$}\color{red}{\ding{55}}\\ Cases $\downarrow$} \\ \hline
\multirow{3}*{GPT3.5} & Self-Reflection & 269& 52 \\
                      & SC-Reflect & 245 & 73 \\
                      &Self-Contrast & \textbf{186} \scriptsize $\downarrow$30.8\% & \textbf{11} \scriptsize $\downarrow$78.9\%  \\ \hline
\multirow{3}*{L-70B} & Self-Reflection & 528& 140 \\
&SC-Reflect & 468 & 127\\
&Self-Contrast & \textbf{401} \scriptsize$\downarrow$24.8\% & \textbf{71} \scriptsize$\downarrow$49.2\%   \\

\bottomrule[0.5pt]
\end{tabular}
\caption{Self-Contrast is evaluated on two cases. }\label{analysis}
\end{table}

%Invalid: Wrong $\rightarrow$ Wrong and Toxic: Right $\rightarrow$ Wrong.

\subsection{Reducing Invalid and Toxic Reflections} 
As mentioned in \Cref{feedback_analyze}, due to overly confident or highly random in the self-evaluate process, vanilla self-reflection contains a large amount of invalid (\color{red}{\ding{55}}\color{black} $\rightarrow$ \color{red}{\ding{55}}\color{black}: 20.3\%) or toxic reflections (\color{green}{\ding{51}}\color{black} $\rightarrow$ \color{red}{\ding{55}}\color{black}: 4\%). Therefore we investigate how Self-Contrast improves these two scenarios on GSM8K. As shown in \Cref{analysis}, we observe that with Self-Contrast, the occurrences of invalid and toxic cases significantly reduced. In particular toxic cases decreased by 78.9\% and invalid cases by 30.8\% using GPT3.5. In contrast, the SC-Reflect does not significantly mitigate either of these scenarios.

The results indicate that through exploration, comparison, and summarization, the uncertainty in the reflection process is greatly reduced, thereby enhancing the error-correction capability of the LLM.

%through exploring, contrasting, and summarizing, thereby achieving more precise and stable reflection.
%These results reveal that Self-Contrast can effectively enhance the self-correction capabilities of LLMs and also significantly reduce toxic reflections caused by uncertainty.
%It indicates that contrasting inconsistent perspectives can effectively alleviate the uncertainty in reflection, particularly in preventing correct responses from being erroneously modified.

\begin{table}[t!]\small
\centering
\setlength\tabcolsep{3pt} % 默认值通常是6pt
\begin{tabular}{ l  c }
\toprule[1pt]
\textbf{\makecell[l]{Strategy}} & \textbf{\makecell[c]{Acc.(\%)}}\\ \hline
Self-Evaluate - \emph{An \textbf{Incorrect} Solution}  &  70.1  \\  
Self-Contrast      \\
- \emph{A \textbf{Correct} and an \textbf{Incorrect} Solutions}  & 83.6 \\
- \emph{Two \textbf{Incorrect} Solutions with \textbf{Similar} Error}  & 70.9 \scriptsize    \\
- \emph{Two \textbf{Incorrect} Solutions with \textbf{Different} Error}  & 75.5 \scriptsize   \\
\bottomrule[0.5pt]
\end{tabular}
\caption{We conduct comparisons across four cases on a subset of GSM8K. LLM self-evaluates or self-contrasts different initial responses and reflects on their results.}\label{two incorrect solution comp}
\end{table}

\subsection{Contrasting Incorrect Solutions is also Instructive}
%Contrasting is Easier than Evaluating 
Self-Contrast inspires reflection by contrasting the differences. An intuitive explanation is that the errors in different responses are dissimilar or randomized, so they can be used to compare with each other and eliminate uncertainties or biases. To verify this, we sample 200 questions from GSM8K, each manually annotated with a correct solution, two incorrect solutions with similar errors (e.g., Error1), and an incorrect solution with a different error (Error2). We design four experiments: 1. Self-evaluate \textbf{one incorrect} solution followed by reflection. 2. Self-Contrast a \textbf{correct} and an \textbf{incorrect} solution. 3. Self-Contrast \textbf{two similar incorrect} solutions. 4. Self-Contrast \textbf{two dissimilar incorrect} solutions. \Cref{two incorrect solution comp} shows that contrasting a correct and an erroneous solution, or contrasting two incorrect solutions with different errors both yield significant enhancements of 13.5\% and 5.4\%. In contrast, comparing two solutions with similar errors does not result in perceptible changes. 

This result aptly explains that the improvement of Self-Contrast stems from contrasting the differences between dissimilar solutions. Therefore, even if candidate solutions are both incorrect, as long as their errors are different, Self-Contrast has the potential to eliminate errors. In other words, Self-Contrast can mitigate the random errors arising from the inherent uncertainty of the LLM.

\begin{figure}[!t] %H为当前位置，!htb为忽略美学标准，htbp为浮动图形
\centering %图片居中
\includegraphics[width=0.46\textwidth]{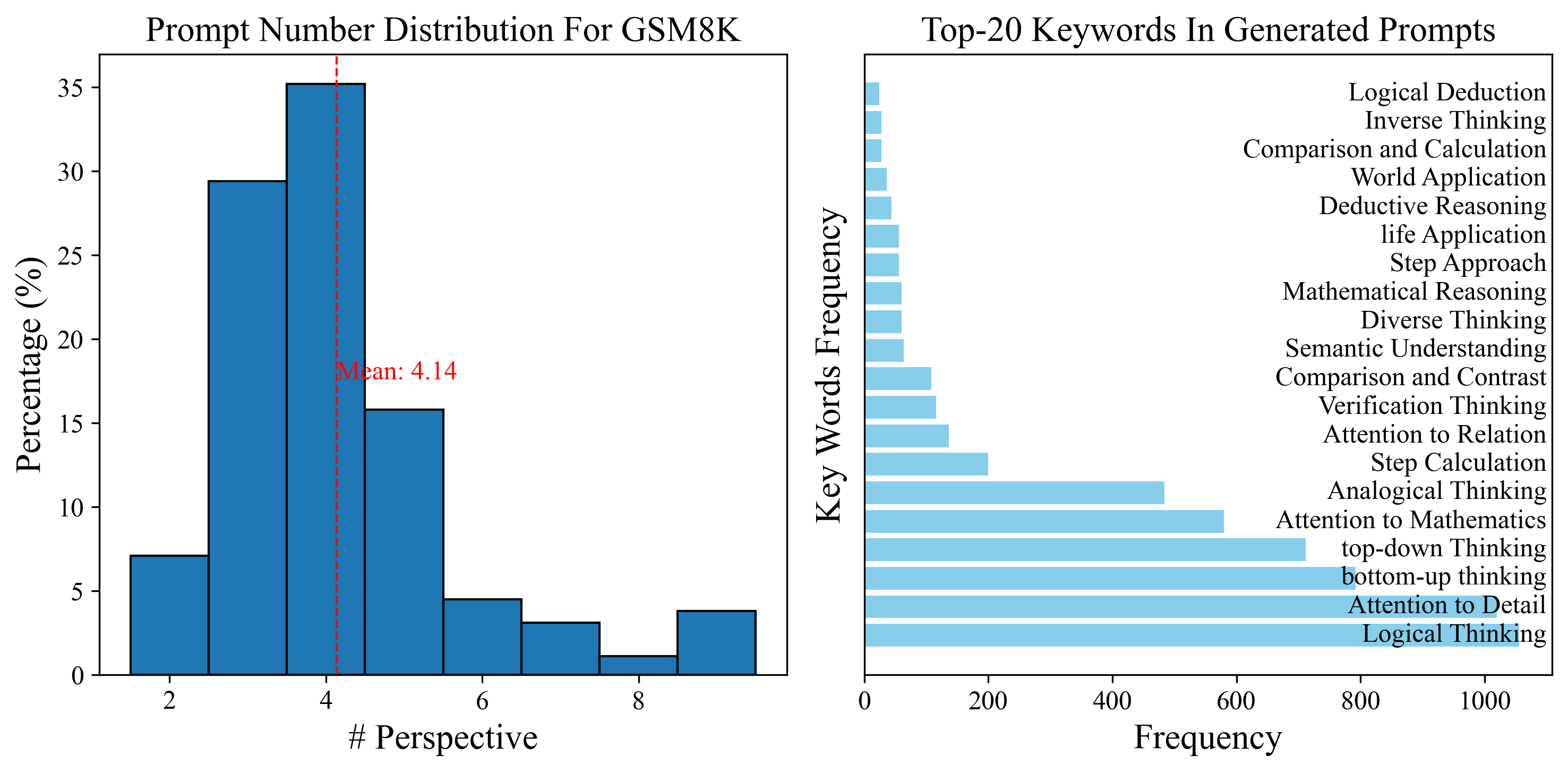} %
\caption{Left: The distribution of the prompt number generated when Self-curated. Right: We visualize the top-20 keywords and frequencies in the prompt name.}\label{self-curated results} %最终文档中希望显示的图片标题 
\end{figure}

\subsection{Diverse Solving Perspectives Maximize Contrast Effect}
Prior analysis indicates that only contrasting dissimilar solutions can foster reflection. Reviewing our strategy, we employ a self-curated prompt process to create multiple solving perspectives ($\mathsection$~\ref{section 3.1}), thereby providing diverse solutions for subsequent comparison. Here, we analyze the distribution of perspectives generated by this process in \Cref{self-curated results}. For most requests, the LLM generates four prompts. We also analyze the frequency of keywords within these perspective's names. For mathematical reasoning, the LLM indeed adaptively designs numerous unique solving perspectives, then generating a variety of results. These dissimilar results can maximize the efficacy of our contrastive strategy.

\section{Discussions}
\textbf{Self-contrast switches the critique objective into a contrastive task}. We transform the self-evaluation into a process of comparing differences, explicitly altering the attention distribution of the LLM. The LLM is required only to identify the differences between two solutions, without judging right or wrong. This process is less influenced by the biases inherent in LLMs, as the objective is contrasting rather than evaluation. Besides, in~\Cref{two incorrect solution comp}, LLMs are instructed to contrast two incorrect solutions with different errors which also improves reflection results.

The results in~\Cref{section2.3} also precisely verify this conjecture. By simply transforming from direct evaluation to contrastive evaluation, we enhance the effectiveness of reflection (75.8 to 77.5 on GSM8K), with more significant results (0.66 to 0.09). In~\Cref{combined_main_result,commonMT_result}, our self-contrast approach achieved more significant improvements.

\textbf{Contrasting results can help LLMs notice overlooked details and biases}. After contrasting the differences between the two solutions, we summarize these differences into a checklist, thereby explicitly prompting the LLM to focus on the logical pitfalls and other issues underlying these differences. This allows LLMs to engage in reflection more clearly and purposefully.

As shown in~\Cref{figure2}, LLM generates different translations for the user’s request: "This plan was shot to death", and "This plan was axed". The former is a rigid translation that fails to grasp the metaphor embedded in the military term. After contrasting two different translations, LLMs believe they should scrutinize the source sentence for metaphors and ensure the translation aligns with the conventions of English expression.

\section{Conclusion}
%这篇文章中，我们针对大模型的反思能力做了详细的调查. 我们发现在缺乏外部反馈时，大模型很难通过自我反思来修正先前response中的错误。分析反思失败的原因，我们发现LLMs无法准确地评估之前的solution。他们自我评估时候生成反馈要么很顽固或者要么很随机，最终影响了自我修正的效果。为了缓解这个问题，我们提出Self-Contrast, 通过对比多个solving视角之间的差异来启发反思，避免了直接评估先前solutions的正确与否。我们的实验表明，Self-Contrast在多个场景下，多种模型下的表现都超过了self-reflection.

We conduct a comprehensive investigation into the inherent reflection capabilities of LLMs. Our findings reveal a notable challenge: in the absence of external feedback, LLMs struggle to correct errors in previous responses on their own. After analyzing their self-evaluate process, we discover that LLMs are unable to accurately evaluate prior solutions and often provide overconfident or inconsistent feedback, which impedes reflection. To mitigate this, we introduce Self-Contrast, a contrastive strategy that inspires reflection by contrasting the differences between multiple perspectives, providing an informative checklist for reflection. Our experiments show that Self-Contrast performs well across a variety of scenarios and with different LLMs.

%It often exhibits an overconfident belief, stubbornly believes its prior solutions as error-free, or alternatively, provides feedback that is seemingly random. This low-quality feedback significantly impedes the effectiveness of self-correction mechanisms. 

\section*{Limitations}
For some smaller-scale LLMs, their instruction-following capability is weaker, hindering their potential to conduct precise comparisons and reflection. In such scenarios, the effectiveness of Self-Contrast might be slightly inferior to ensemble strategies. For instance, the performance of Self-Contrast with Llama2-7B is marginally lower than self-consistency. A viable approach is to utilize an external tool to compare differences between multiple perspectives, rather than LLM itself. For instance, we explore utilizing sequences comparison library \emph{difflib}\footnote{\url{https://docs.python.org/3/library/difflib.html}} to contrast two generated equations (e.g., differ.compare(a+b$\div$c, a-b$\div$c)) or some rule-based strategy to compare two responses. It can provide us with more accurate and flexible comparisons at different granularity (e.g., character level). We leave this as future work.

\section*{Acknowledgments}
This work is supported by the National Natural Science Foundation of China (No. 62376245), the Key Research and Development Program of Zhejiang Province, China (No. 2024C01034),  the Fundamental Research Funds for the Central Universities, and National Key Research and Development Project of China (No. 2018AAA0101900).

\iffalse
\section{Analysis}
\begin{table}[t!]\small
\centering
\begin{tabular}{ l | c  c  c  c  c   }
\toprule[1pt]
 Sub-Set& \makecell[c]{Valid\\\scriptsize W-R} & \makecell[c]{Failed\\\scriptsize W-W} & \makecell[c]{Toxic\\\scriptsize R-W} & \makecell[c]{Easy\\\scriptsize R-R} & Overall\\ \hline
\makecell[l]{\# Number} & 48 & 269 & 52 & 950 & 1319 \\ \hline
Cot-Prompt & 0& 0& 52& 950 &   1002\\
Self-Reflection & 48& 0& 0& 950 &  998\\
SC-Reflect & 31 & 17 & 5   & 946  &999\\
Self-Contrast & 44 & 93 & 30 & 945  &1112\\
\bottomrule[0.5pt]
\end{tabular}
\caption{Analysis}\label{indirect feedback}
\end{table}

- 对抗低质量的initial response：用LLama生成多个视角的initial response, 然后用gpt35做对比和graph reflection，看精度。和标准的self-contrast比，和带有噪声的self-consistency比

分析multiple视角的个数分布，以及个数为n划分子集 计算对应的那些样本的反思前后精度变化

graph reflection反思的轮次和精度的关系initial
\fi

%We guide LLMs to identify a series of auxiliary tasks which may play a positive role in reasoning.

\bibliography{custom}
\bibliographystyle{acl_natbib}

\appendix

\clearpage
\renewcommand\thefigure{\Alph{section}\arabic{figure}}    
\setcounter{figure}{0}    
\renewcommand\thetable{\Alph{section}\arabic{table}}    
\setcounter{table}{0}   

\noindent\textbf{Appendix}

\section{Complementary Experiments}
\subsection{Detail For Manual Feedback Evaluation}\label{Detail For Manual evaluation}
We provide the details of manual evaluation in~\Cref{feedback_analyze}. Specifically, we categorize reflection results into four categories: Invalid (wrong -> wrong), Valid (wrong -> right), Toxic (right -> wrong), and Other (right -> right) based on their answer. Subsequently, we manually assess the quality of feedback within each reflection category. For instance, for a Toxic reflection case, we devise ten self-evaluation prompts, prompting LLMs to conduct a self-evaluation on their initial response, generating ten feedbacks. These feedbacks are manually checked for correctness and consistency. The criteria for classification are as follows:
\begin{itemize}
    \item If more than seven feedbacks accurately identify the errors in the initial response, we classify it into the first category: \textcolor{blue}{I. Accurately Identifies Errors}. Please note that the feedback only needs to accurately identify the errors without necessarily correcting them.
    
    \item If more than seven feedbacks indicate that the initial response has no errors, we categorize it into the fourth category: \textcolor{blue}{IV. Overconfidence - No Revision Required.}
    
    \item If, among the ten feedbacks, there are more than three different opinions, e.g., the first feedback suggesting there are no errors, another identifying error-1, and yet another pointing out error-2.... For this scenario, we classify it into the third category: \textcolor{blue}{III. Cannot Output Consistent Feedback}. In this scenario, self-evaluation exhibits significant randomness.
    \item The remaining cases are classified into the second category: \textcolor{blue}{II. Stubbornly Offers Erroneous Feedback}.

\end{itemize}

The entire human evaluation process is conducted by two senior PhD students. One is responsible for categorization, while the other verifies the categorization again.

\subsection{LLM is More Likely to Trust Previous Response}
We investigate whether LLMs are prone to uncritically trusting previous responses during reflection, rather than meticulously examining and rectifying errors. Typically, self-reflection often contains three stages, i.e., initial response, self-evaluate, and revision stage. We employ different LLMs to provide a poorer quality response as the initial response for the subsequent two stages. We observe whether this affects the results of the reflection, e.g., we replace gpt3.5$\rightarrow$gpt3.5$\rightarrow$gpt3.5 with Llama-2-70b$\rightarrow$gpt3.5$\rightarrow$gpt3.5. If LLMs tend to place undue trust in prior responses, the efficacy of the final reflective process will be adversely impacted.

\begin{figure}[!h] %H为当前位置，!htb为忽略美学标准，htbp为浮动图形
\centering %图片居中
\includegraphics[width=0.5\textwidth]{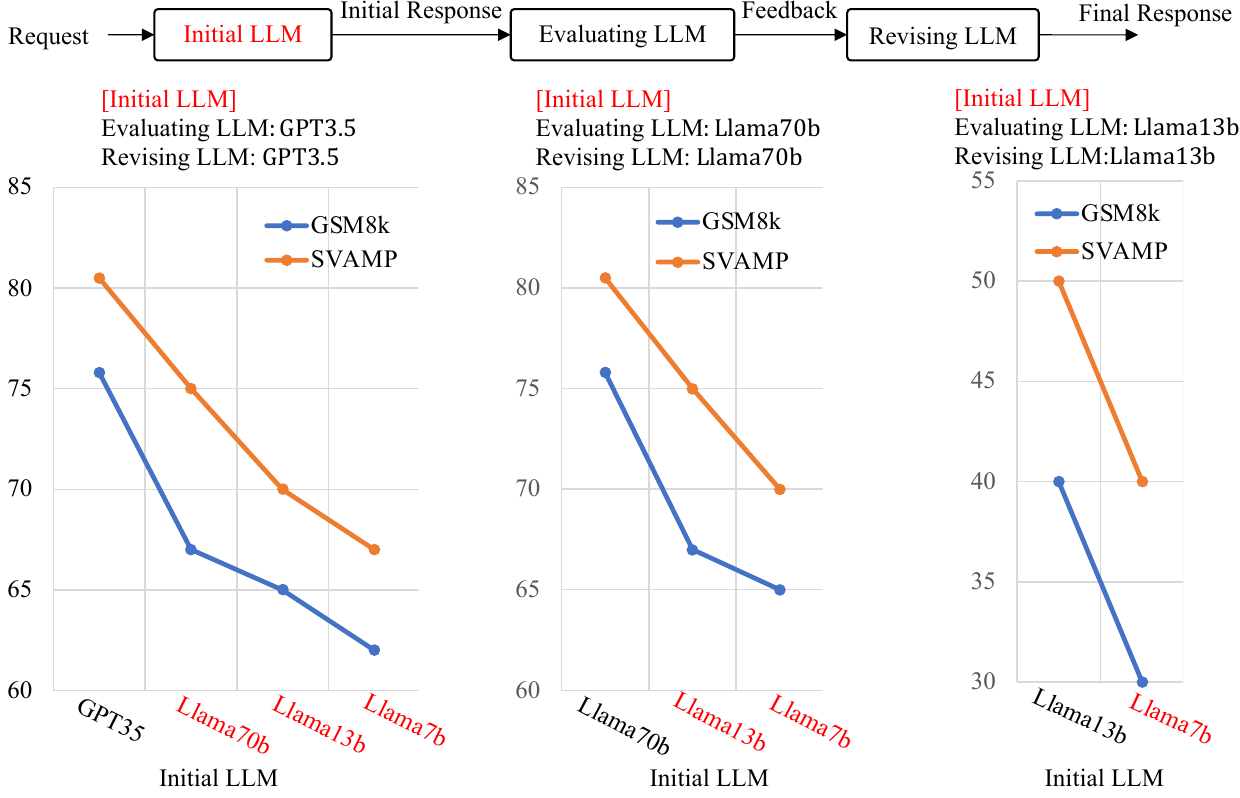} %
\caption{The Reflection Accuracy with Different LLM for Initial Response. \emph{Left}: different LLMs provide initial responses when GPT3.5 is utilized for Evaluation and Revision. \emph{Center}: different LLMs provide initial responses when Llama2-70B is utilized for Evaluation and Revision. \emph{Right}: different LLMs provide initial responses when Llama2-13B is utilized for Evaluation and Revision. The results indicate that LLMs are easily influenced during reflection. LLM is predisposed to trust previous responses over diligently examining and correcting errors.}\label{weak_llm_figure} 
\end{figure}

However, as shown in \Cref{weak_llm_figure}, the reflection results are severely impacted by the quality of the initial response. E.g., compared with using gpt3.5 for three phases, Llama2-70b$\rightarrow$gpt3.5$\rightarrow$gpt3.5 exhibits a marked decrease (-8.4\% for GSM8K). Furthermore, we also observe the weaker the LLM replaced, the poorer the performance after reflection. It suggests that LLMs tend to trust the initial solution rather than detect and revise the errors during the self-evaluate phase. 

\begin{figure}[t!] %H为当前位置，!htb为忽略美学标准，htbp为浮动图形
\centering %图片居中
\includegraphics[width=0.46\textwidth]{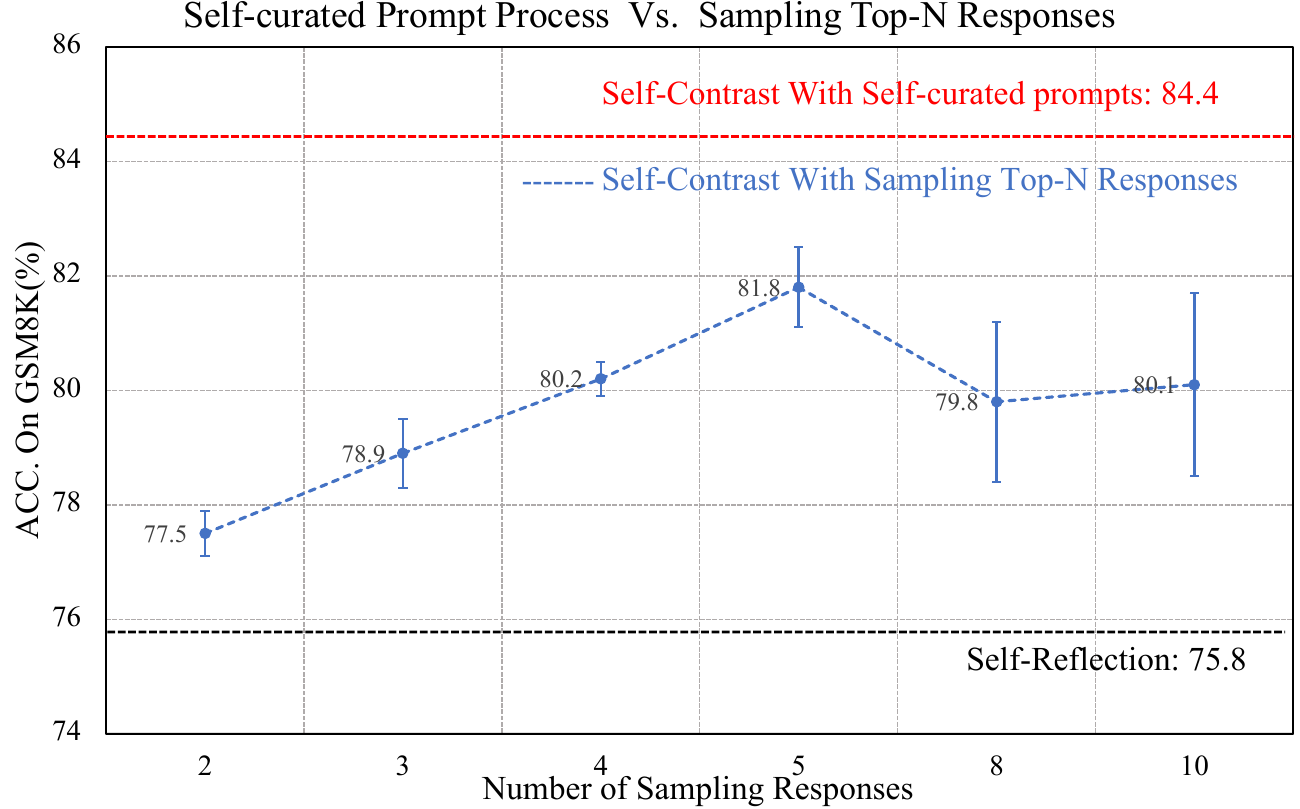} %
\caption{We replace the self-curated prompt process with a simple strategy: directly sampling top-n responses for contrast. We observe that as N increases, the performance also improves, yet it still remains lower than self-contrast with the self-curated prompts. All results are conducted on GSM8K using GPT-3.5.} %最终文档中希望显示的图片标题
\label{Fig.sample_top-n} %用于文内引用的标签
\end{figure}

\begin{table}[t!]\small
\centering
\setlength\tabcolsep{3pt} % 默认值通常是6pt
\begin{tabular}{ l  c   c}
\toprule[1pt]
\textbf{\makecell[l]{Strategy}} & \textbf{\makecell[c]{GSM8K}} & \textbf{\makecell[c]{CommonMT}}\\ \hline
Self-reflection  & 75.8 & 69.3 \\ 
Self-Contrast    & 84.4  & 70.7 \\
- \emph{w/o Checklist Generation}  & 80.9 \scriptsize $\downarrow$3.5  &70.6 \scriptsize $\downarrow$0.1  \\
Selecting Strategies     \\
- \emph{Random Selecting}  & 76.4 \scriptsize $\downarrow$8  &69.5 \scriptsize $\downarrow$1.2  \\
- \emph{Clustering + Random Selecting}  & 81.2 \scriptsize $\downarrow$3.2  &69.7 \scriptsize $\downarrow$1.0  \\
- \emph{Clustering + LLM Selecting}  & 82.6 \scriptsize $\downarrow$1.8  &69.9 \scriptsize $\downarrow$0.8  \\
- \emph{Clustering + Negative Perspective}  & 83.9 \scriptsize $\downarrow$0.5  &70.8 \scriptsize $\uparrow$0.1  \\

\bottomrule[0.5pt]
\end{tabular}
\caption{We eliminate the checklist generation process, instructing the LLM to directly reflect on the differences from contrasting multiple perspectives. Besides, we also analyze the impact of different selecting strategies.}\label{reflection on differences}
\end{table}

%- \emph{w/o Negative Perspective}  & 83.0 \scriptsize $\downarrow$1.4 & 69.9 \scriptsize $\downarrow$0.8 \\
%Then, we remove the Negative Perspective and observe its results. All results are conducted on GSM8K using GPT-3.5

\subsection{Self-Evaluate Vs. Self-Contrast}
\begin{figure}[!h] %H为当前位置，!htb为忽略美学标准，htbp为浮动图形
\centering %图片居中
\includegraphics[width=0.46\textwidth]{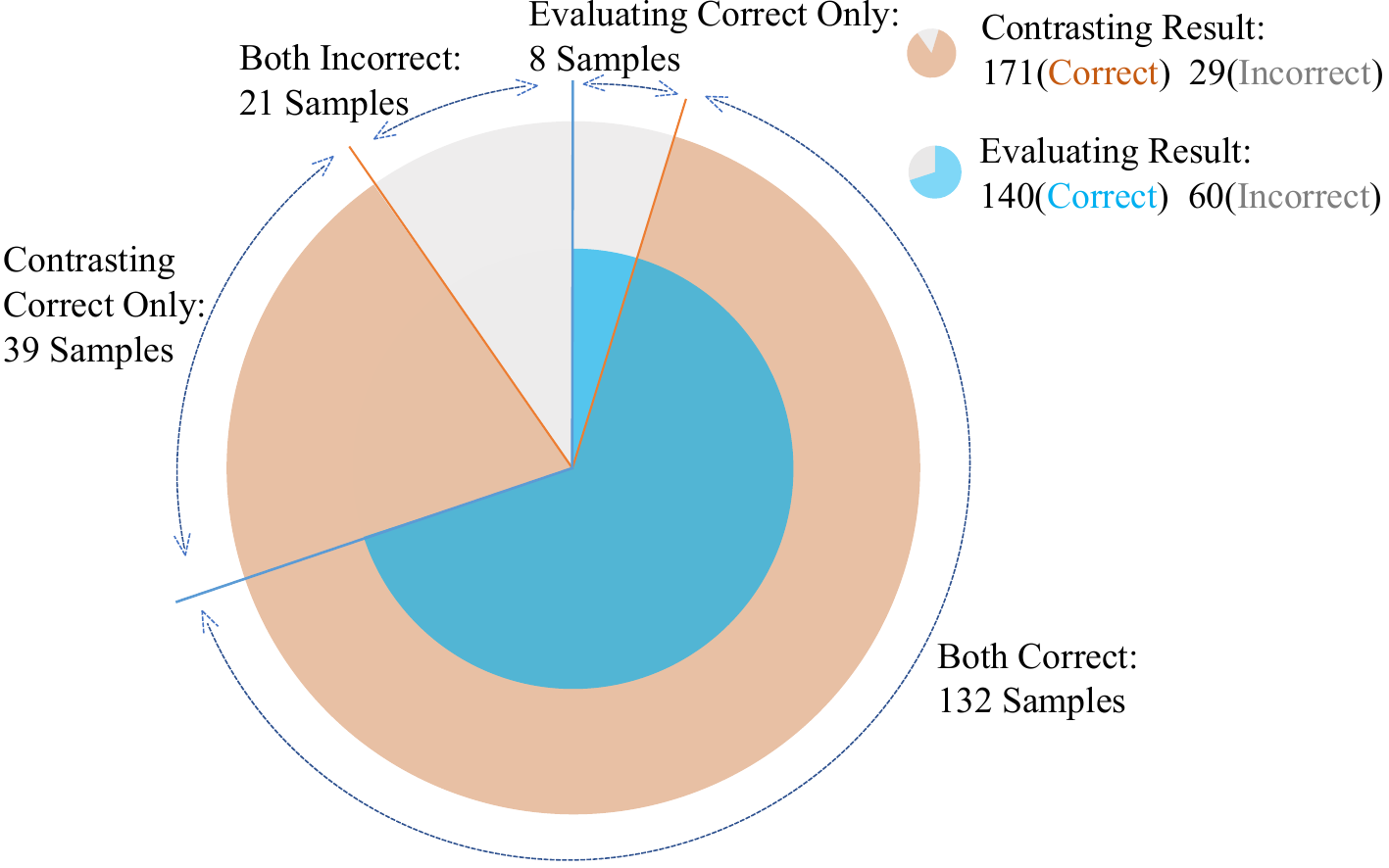} %
\caption{We compare the results of the \emph{Evaluating} and \emph{Contrasting} using two pie charts. It shows \emph{Contrasting} is more accurate and stable than direct \emph{Evaluating}.} %最终文档中希望显示的图片标题
\label{Fig.analyze} %用于文内引用的标签 
\end{figure}

Self-Contrast inspires reflection by contrasting the differences, rather than evaluating directly. The underlying assumption is contrast is more accurate and stable than direct evaluation for LLM. To validate this, we conduct an experiment using 200 samples from GSM8K, each containing a correct and an incorrect solution. We design two tasks: Taks 1: Contrasting two solutions. Task 2: Evaluating the incorrect solution. We manually check the results of two tasks, i.e., whether LLM can perform contrast or evaluate correctly. As shown in \Cref{Fig.analyze}, we observe contrasting is more accurate than direct evaluating (171 correct Vs. 140 incorrect).

Further, we divide all samples into four cases: 1. both tasks are correct. 2. Contrasting: correct, Evaluating: wrong. 3. Contrasting: wrong, Evaluating: correct. 4. Both are wrong. In \Cref{Fig.analyze}, the results show that when LLM can correctly evaluate a solution, it is often able to contrast correctly, with few exceptions (only 8 samples for Evaluating Correct Only). Notably, in 39 cases, the LLM fails in direct evaluation but succeeds in contrast. These results indicate that contrasting two solutions is more accurate and stable than direct evaluation, leading to more reliable results.

\subsection{Ablation Study For Selection Strategy}
As introduced in \Cref{section 3.2}, we cluster multiple responses generated by the self-curated process and then select the cluster center from each category for contrast. We design four different selection strategies. 1) \emph{Random Selecting}: We randomly choose $K$ responses from all candidates. 2) \emph{Clustering + Random Selecting}: We first cluster all responses into $k$ categories, then randomly select one from each category. 3) \emph{Clustering + LLM Selecting}: Similarly, we first cluster all responses into $k$ categories, then instruct the LLM to choose a potentially correct response from each category. 4) \emph{Clustering + Negative Perspective}: We first instruct the LLM to consider what are common errors for the user request. Then LLM should intentionally generate an imperfect solution based on these common errors. Finally, we instruct the LLM to select one response from each category that is least similar to the intentionally generated imperfect solution. As shown in \Cref{reflection on differences}, we observe that compared to Self-Contrast, the performance of several selection strategies experiences a certain degree of decline.

\section{Experiments Details} 
\subsection{Benchmarks} \label{benchmarks}
\textbf{Mathematical Reasoning:} We leverage multiple datasets with different complexity and languages, including GSM8K~\citep{Cobbe2021TrainingVT}, SVAMP~\citep{patel-etal-2021-nlp} as benchmarks to evaluate performance. Notably, GSM8K presents higher levels of difficulty, encompassing complex mathematical operations, while SVAMP is slightly simpler and consists of combinations of addition, subtraction, multiplication, and division.

\textbf{Creative Translation} In addition to mathematical reasoning tasks, we introduce a generation task: creative translation. We utilize the CommonMT~\cite{he-etal-2020-box}, which includes a vast body of Chinese-to-English pair examples. Unlike conventional translations, most samples contain non-standard expressions such as idioms and metaphors, necessitating an understanding of local cultural and linguistic habits for accurate translation. Following the Multi-agent debate~\citep{Liang2023EncouragingDT}, we adopt the samples with "hard" categories from CommonMT as testing benchmarks.

\subsection{Other Details} \label{other details}
In the Self-Curated Prompt phase, we limit LLMs to design at least two different prompts and a maximum of nine prompts for each request. In selecting stage (\Cref{section 3.2}), we set $k$ to 3, which means that all perspective results are divided into three categories, and then we select a result from each category. We instruct LLM sequentially output comparisons among three results, subsequently synthesizing these differences into a comprehensive checklist in a single pass, eliminating the need for multiple prompts. Besides, due to the diversity of translation tasks, we also introduce a negative perspective for translation. Specifically, we instruct LLMs to consider what common errors might be made for the user request, then actively adopt a careless persona to generate an incorrect response with some common mistakes. The result of this negative perspective serves as a negative demonstration for subsequent selection and reflection. 

\section{Baseline Prompts}\label{baseline_prompt}
\lstset{
  breaklines=true, % 设置自动换行
  basicstyle=\ttfamily\small, % 设置基本样式为等宽字体
  breakindent=0pt,
  escapeinside={(*@}{@*)}, % 设置 escapeinside 选项
}

\iffalse
2: Help me work through following math question step-by-step. The question is {question}.
3: Please Solve the math question step-by-step. The question is {question}.
4: Now please carefully solve the following math question step-by-step. The question is {question}.
5: Please show detailed steps to solve this math problem. The question is {question}.
2: Please provide a detailed translation of this Chinese paragraph into English. The Chinese sentence is {sentence}.
3: Now, let's carefully translate the following Chinese paragraph into English step by step. The Chinese sentence is {sentence}.
4: Please translate Chinese into English accurately. The Chinese sentence is {sentence}.
5: What is the most accurate English translation of the following Chinese sentence? The Chinese sentence is {sentence}.
\fi

\textbf{Standard Prompt} We use a simple prompt for CoT Prompt and self-consistency baselines. For each experiment, we run 10 times and averaged their results.

\begin{lstlisting}
(*@\textbf{Math Reasoning}@*): You are a math teacher. Let us solve the math question step by step. The question is {(*@\color{blue}{input}@*)}.
\end{lstlisting}

\begin{lstlisting}
(*@\textbf{Creative Translation}@*): You are an expert translator, please translate Chinese into English accurately. The Chinese sentence is {(*@\color{blue}{input}@*)}.
\end{lstlisting}

\noindent\textbf{Reflection Prompts} We designed 10 prompts for the self-reflection baseline. Each experiment follows \emph{Initial response-Evaluation-Revision} pattern. The prompt for the \emph{Initial response} remains consistent with previous experiments (Standard Prompt).

\begin{lstlisting}
1:
(*@\textbf{Evaluation}@*): Please carefully examine the previous responses for correctness, and provide detailed feedback.

(*@\textbf{Revision}@*): Please refine the previous response based on the feedback.
2:
(*@\textbf{Evaluation}@*): Please review your previous responses for any errors, and provide detailed feedback. 

(*@\textbf{Revision}@*): Please refine the previous response based on the feedback. If there are no questions, you can repeat the previous solution
3:
(*@\textbf{Evaluation}@*): Do you think the previous response is correct or not, and if not please point out where is wrong.

(*@\textbf{Revision}@*): Please refine the previous response.
4:
(*@\textbf{Evaluation}@*): Please carefully evaluate the quality of the previous response and point out if you feel something is not appropriate

(*@\textbf{Revision}@*): Please carefully consider the comments in the feedback and re-generate the answer.
5:
(*@\textbf{Evaluation}@*): Please double-check the previous response for any errors. If there are any errors, please point them out.

(*@\textbf{Revision}@*): Please read the feedback carefully, and improve your answer.
6:
(*@\textbf{Evaluation}@*): There may have been some mistakes with your previous response, so please double-check and find out the mistake. If you think there are no errors at all, please just reply, "Exactly correct".

(*@\textbf{Revision}@*): Please refine your response. If you think it's acceptable, then just repeat your last response.
7:
(*@\textbf{Evaluation}@*): Please check that your previous response matches the question. Please point out if it does not fit

(*@\textbf{Revision}@*): Please refine your response based on the feedback. If the feedback points out something that is not perfect please fix it!
8:
(*@\textbf{Evaluation}@*): Please consider whether your response addresses the problem. If not or if there is an error please point it out

(*@\textbf{Revision}@*): Please reflect based on the feedback and improve your response.
9:
(*@\textbf{Evaluation}@*): Please assess in detail whether your previous response solves the problem and provide feedback.

(*@\textbf{Revision}@*): Please refine your response based on the feedback.
10:
(*@\textbf{Evaluation}@*): Please check your previous response for correctness and whether it can be further enhanced.

(*@\textbf{Revision}@*): Please further refine your response based on the feedback. If you don't feel it is necessary then restate the previous response
\end{lstlisting}

\section{Our Prompt} \label{our prompts}
\subsection{Prompt for Self-Curated Process} \label{prompt for self-curated}
Different requests may require some unique solving perspectives. We design a self-curated prompts process, enabling LLMs to design their prompts based on specific user requests. The prompt for the self-curated process is as follows:
\begin{lstlisting}
(*@\textbf{Translation Task:}@*)
You are a translation specialist who specializes in translating from diverse perspectives. Given a Chinese source sentence, you need to carefully analyze the source sentence and dynamically generate several useful prompt instructions. These prompt instructions should be diverse and also relevant to the source sentence. These prompt instructions are used to guide the language model to think in different ways, attention to different emphases, and reason from different perspectives for a more accurate translation.

For instance, you can design different translation styles, different expressions of emotion, different emphases, and different tones for input sentences in prompt instruction. Besides you can create different knowledge backgrounds, identities, personalities, different concerns, etc for more relevant translation.

Here are some guidance rules for Prompt Generation:
1. Tone Requirement: Please generate prompt instructions in the third person.  
2. Content Requirement: Each prompt instruction should be different, and include at least three parts: translation styles, attention emphasis, and tones and emotion design. Please do not state them separately.
3. Number Requirement: Dynamically generate the most valuable 2 to 9 prompt instructions based on the input Chinese source sentences.
4. Format Requirement: Each prompt instruction should start with ###.
5. Others: Prompt should focus on translation. So don't ask any other irrelevant questions in the prompt.



Here is an example:
The input Chinese sentence is: (*@\begin{CJK*}{UTF8}{gkai}他想拉同村的干部一起下水去贩毒\end{CJK*}@*). Please generate the most suitable prompts. 
Output:
Literal Perspective: ###You are a meticulous translator, proficient in direct translation, and highly focused on specifics. Your translation approach prioritizes precise replication of the original text's expression.

Liberal Perspective: ###You are an inventive translator, characterized by a dynamic and liberal translation approach, often reimagining the original text's meaning in your own linguistic style.

The input Chinese sentence is  {(*@\color{blue}{input}@*)}. Please generate the most suitable prompts: 
\end{lstlisting}

\begin{lstlisting}
(*@\textbf{Reasoning Task:}@*)
You are a math specialist who specializes in math solving from diverse perspectives. Given a math question, you need to carefully analyze the question and dynamically generate several useful prompt instructions. These prompt instructions should be diverse and also useful for math-solving. These prompt instructions are used to guide the language model to think in different ways, attention to different emphases, and reason from different perspectives for more accurate math solving.

For instance, you can adopt multi-faceted thinking (logical thinking, lateral thinking, analogical thinking, etc.), different reasoning perspectives (e.g., top-down, bottom-up, step-by-step), and different emphases of concern, (entity words, numbers, units, percentages, math knowledge, etc) for input question in prompt instruction.

Here are some guidance rules for Prompt Generation:
1. Tone Requirement: Please generate prompt instructions in the third person.  
2. Content Requirement: Each prompt instruction should adopt a different way of thinking, or focus on a different perspective, or different emphases to solve the question.
3. Number Requirement: Dynamically generate the most valuable 2 to 9 prompt instructions based on the input math question.
4. Format Requirement: Each prompt instruction should start with ### and end with @@@.
5. Others: Prompt instructions should focus on math solving. So don't ask any other irrelevant questions in the prompt.

Here is an example: The math question is: Mark works at his job for 8 hours a day for 5 days a week.  He used to make $10 an hour but they raised his pay by $2 per hour.  How much does he make a week?

Output: 
bottom-up perspective: ###As a mathematician, you have to solve the given problem from a bottom-up perspective. Please focus initially on the foundational elements of the problem. Start with the simplest parts and their interrelations. Progressively build upon these foundational components, joining them together until a complete solution emerges

The input math question is {(*@\color{blue}{input}@*)}. Please generate the most suitable prompts: 
\end{lstlisting}

\iffalse
\subsection{Prompt for Incorrect Perspective}\label{Incorrect Perspective}
\begin{lstlisting}
(*@\textbf{Translation Task:}@*)
Please intentionally generate an English translation that is imperfect or erroneous, which may contain some common errors made by a careless person, e.g., forgetting certain details and translating inappropriately. Such an incorrect response should be distinctly different from the right response. The source sentence is {(*@\color{blue}{input}@*)}. 
(*@\textbf{Reasoning Task:}@*)
Please intentionally generate a math-solving process that is imperfect or erroneous, which may contain some common errors made by a careless person, e.g., forgetting certain details or reasoning incorrectly. Such an incorrect response should be distinctly different from the right response. The question is {(*@\color{blue}{input}@*)}. 
\end{lstlisting}

\fi

\subsection{Prompt for Contrasting Process} \label{edge generating}
\begin{lstlisting}
(*@\textbf{Translation Task:}@*)
You are an expert translator. Given some candidate English translations for a Chinese source sentence, you should carefully compare the difference between each two translations in terms of semantics, syntax, words (e.g., nouns and verbs), and any other aspects. 

When you compare, you need to consider the following questions:
1: Are there differences between the two translations?
2: Where are the differences?
3: What causes these differences?

After contrasting, you should generate a checklist based on these differences between candidate translations. You should carefully consider each discrepancy and the reasons behind it, summarizing them into a few checking instructions in the checklist. This checklist can guide others to re-examine the input sentence and these candidate translations to eliminate these discrepancies. 


Input Format: 
The Chinese sentence is {(*@\color{blue}{Chinese sentence}@*)}. 
All Results: {(*@\color{blue}{Result1}@*)},{(*@\color{blue}{Result2}@*)}, {(*@\color{blue}{Result3}@*)},.... 

Output Format:
For Result1 and Result2: {(*@\color{blue}{Difference1}@*)}.
For Result1 and Result3: {(*@\color{blue}{Difference2}@*)}
For Result2 and Result3: {(*@\color{blue}{Difference3}@*)}
Checklist: {(*@\color{blue}{Directive1, Directive2, ...}@*)}
....

(*@\textbf{Reasoning Task:}@*)
You are a math specialist who specializes in math solving. Given some candidate solutions for a math question, you should carefully compare the difference for each two solutions in their solving steps. 

When you compare, you need to consider the following questions:
1: Are the two solutions have different final answers and mathematical expressions?
2: Where are the differences in their solution steps and mathematical expressions?
3: Why are the answers of the two solutions different?

After contrasting, you should generate a checklist based on these differences between candidate solutions. You should carefully consider each discrepancy and the reasons behind it, summarizing them into a few checking instructions in the checklist. This checklist can guide others to re-examine the input question and these candidate solutions to eliminate these discrepancies. 


Input Format:
The math question is {(*@\color{blue}{Question}@*)}. 
All solutions: {(*@\color{blue}{Solution1}@*)}, {(*@\color{blue}{Solution2}@*)}, {(*@\color{blue}{Solution3}@*)}, .... 

Output Format:
For Solution1 and Solution2: {(*@\color{blue}{Difference1}@*)}
For Solution1 and Solution3: {(*@\color{blue}{Difference2}@*)}
For Solution2 and Solution3: {(*@\color{blue}{Difference3}@*)}
Checklist: {(*@\color{blue}{Directive1, Directive2, ...}@*)}

\end{lstlisting}

\subsection{Prompt For Reflection Stage} \label{prompt format}
We record all candidate responses, their differences, and the checklist in a JSON format. The whole prompt for math reasoning is as follows:

\begin{lstlisting}
(*@\textbf{Reflection Instruction:}@*)
Given a math question, multiple inconsistent solutions, their differences in their solving processes, and a checklist. You should revise the inconsistent solving step for each solution, eliminate the differences, and output a new solving process for each solution.

Guidance Rules for Reflection:

1. Please check carefully according to the requirements on the checklist. It helps you to resolve conflicts between different solutions.
2. When you finish revising inconsistent solutions, please ensure all revised solutions should have the same answer. If not, please revise again until all inconsistencies are removed, and all candidates are consistent.
3. Please output all revised solutions in JSON format as input, without any other text.


The math question is {(*@\color{blue}{question}@*)}. 
The candidate solutions and their discrepancy are as follows: 
{
  "Candidate": {
        "result_1": {
              "answer": "{(*@\color{blue}{answer1}@*)}",
              "solution": "{(*@\color{blue}{solution1}@*)}"},
        "result_2": {
              "answer": "{(*@\color{blue}{answer2}@*)}",
              "solution": "{(*@\color{blue}{solution2}@*)}"},
        "result_3": {
              "answer": "{(*@\color{blue}{answer3}@*)}",
              "solution": "{(*@\color{blue}{solution3}@*)}"},
        ....
  },
  "Discrepancy": {
        "difference_1_2": {
              "source": "result_1",
              "target": "result_2",
              "relation": "{(*@\color{blue}{difference}@*)}" },
        "difference_1_3": {
              "source": "result_1",
              "target": "result_3",
              "relation": "{(*@\color{blue}{difference}@*)}" },
        "difference_2_3": {
              "source": "result_2",
              "target": "result_3",
              "relation": "{(*@\color{blue}{difference}@*)}" },
        ....
  }
}
(*@\textbf{Checklist:}@*) {(*@\color{blue}{Directive1}@*), (*@\color{blue}{Directive2}@*),....}
Please revise each inconsistent solution.

\end{lstlisting}

\iffalse
\begin{table}[h!]
\small
\centering
\begin{tabular}{c|c|c}
\toprule[1pt]
\textbf{} & \textbf{Fixed} & \textbf{Dynamic} \\ \hline
\textbf{Singular} & \makecell[l]{CoT\citeyearpar{wei2022chain},PAL\citeyearpar{gao2022pal}\\Zero-shot-CoT\citeyearpar{wei2022finetuned}}& \makecell{Expert\citeyearpar{Xu2023ExpertPromptingIL}} \\ \hline
\textbf{Multiple} & \makecell[l]{Multi-persona\citeyearpar{2023UnleashingCS}\\AgentDebate\citeyearpar{Liang2023EncouragingDT,Chan2023ChatEvalTB}\\Collaboration\citeyearpar{Dong2023SelfcollaborationCG}\\  AgentVerse\citeyearpar{Chen2023AgentVerseFM}\\MetaGPT\citeyearpar{Hong2023MetaGPTMP}\\AutoGen\citeyearpar{Wu2023AutoGenEN} 
 }&\makecell[c]{Self-Curated\\Prompts}\\ 
\bottomrule[0.5pt]
\end{tabular}
\caption{We consider existing methods from two aspects: fixed / dynamic and singular / multiple perspectives.}
\label{prompting strategy}
\end{table}
\fi

\section{Related Works}
\subsection{Self-correction Ability of LLM} 
Recently, one exciting discovery is that LLMs appear to possess advanced cognitive intelligence: self-correction, where LLMs can refine their previous responses based on feedback~\citep{Shinn2023ReflexionAA, Madaan2023SelfRefineIR, Paul2023REFINERRF}. This capacity endows LLMs to harness external feedback, or even self-evaluated feedback to refine the prior responses~\citep{Welleck2022GeneratingSB, Kadavath2022LanguageM, Chen2023TeachingLL, Kim2023LanguageMC, Xi2023SelfPolishER, Ganguli2023TheCF, Pan2023AutomaticallyCL, Nathani2023MAFMF}. This capacity, particularly when it is solely reliant on inherent reflection, has generated significant interest in the academic community. It appears that a simple iterative prompt strategy could facilitate self-correction in an LLM-based system. However, recent studies~\citep{Huang2023LargeLM, Stechly2023GPT4DK, Liang2023EncouragingDT, Valmeekam2023CanLL} have cast doubt on LLM's inherent reflection capability. Their research indicates that without external feedback, LLMs have difficulties in amending prior responses.

\subsection{Prompting for Better Problem-Solving}
Drawing on cognitive science, human reasoning involves two different reasoning patterns: breadth reasoning, i.e., exploring various reasoning perspectives, and depth reasoning, which involves continually refining ideas and minimizing errors. Based on this concept, we can view previous prompting strategies as either breadth or depth reasoning. Self-consistency and some contemporaneous works~\citep{Wang2023Self, Huang2022LargeLM, Yoran2023AnsweringQB, Jain2023SelfconsistencyFO, Chen2023UniversalSF} mimic breadth reasoning by sampling diverse reasoning processes and voting the final answer, while self-reflection, abstraction reasoning strategies \citep{Shinn2023ReflexionAA, Madaan2023SelfRefineIR, Paul2023REFINERRF, Zheng2023ProgressiveHintPI, Wang2023PlanandSolvePI, Yoran2023AnsweringQB, Zheng2023TakeAS, Xu2023ReReadingIR,shridhar2023art} represent depth reasoning, refining reasoning through iterative prompting strategy. Except for these, Self-Verification~\citep{Weng2022LargeLM} designs a reverse generation from the answer to given conditions, which is widely used in machine translation~\citep{edunov2018understanding}. \citet{Cohen2023LMVL, Mndler2023SelfcontradictoryHO} propose a method for detecting self-contradictions or factual errors in responses to enhance quality. However, our Self-Contrast combines both breadth and depth reasoning. It creates multiple perspectives to enhance the breadth of reasoning and also reflects on the differences for better depth reasoning, offering more reliable problem-solving.

\subsection{Agent-based Methods}
Recent studies~\citep{li2023camel, Deshpande2023ToxicityIC, Xu2023ExpertPromptingIL, Du2023ImprovingFA, Xiong2023ExaminingIO} have found that when an LLM is assigned a specific role personas, it can generate higher-quality responses. This suggests that LLMs are powerful enough, and the appropriate prompt can elicit this capability. Moreover, recent works~\citep{Wang2023UnleashingCS, Fu2023ImprovingLM, Liang2023EncouragingDT, Schick2022PEERAC, Dong2023SelfcollaborationCG, Park2023GenerativeAI, Liu2023DynamicLN} have utilized a multi-role dialogue to collaborate or debate with each other for a more comprehensive response. Furthermore, some studies~\citep{Chen2023AgentVerseFM, Chan2023ChatEvalTB, Huang2023EnhancingLL, Chen2023AutoAgentsAF, Hong2023MetaGPTMP, Wu2023AutoGenEN} have integrated this concept with complex tasks such as code generation by decomposing a complex task into several sub-tasks and employing multiple agents with different identities for each sub-task.  However, most agent-based approaches necessitate careful manual design of each agent's role and pattern of interaction. Our approach, in contrast, does not require pre-defined agents' roles and numbers by humans, as it is entirely designed by the LLMs based on the user request, offering greater flexibility.

\subsection{Learning Mathematical Reasoning} 
Mathematical reasoning is the key to achieving embodied intelligence~\citep{Zhang2021LearningTN, ijcai2022p654}. In recent years, mathematical reasoning has become a significant benchmark \citep{Cobbe2021TrainingVT, Hendrycks2021MeasuringMP} to evaluate the capabilities of artificial intelligence models. Within the paradigm of supervised learning, a vast amount of research~\citep{xie2019goal, patel-etal-2021-nlp, jie-etal-2022-learning, zhang-etal-2022-multi-view, zhang-etal-2023-expression} has been dedicated to translating human language into mathematical equations. In the era of LLMs, the advent of Chain-of-Thought and other prompting strategies have notably augmented the reasoning capabilities~\citep{zhu-etal-2023-solving, Yuan2023HowWD, Frieder2023MathematicalCO, Zhou2022LeasttoMostPE}. 

\textbf{Prompting Method} PAL and Program-of-Thoughts~\cite {gao2022pal, Chen2022ProgramOT} separate the computation and reasoning process using code as the intermediate process. Mathprompter, Auto-Model~\citep{Imani2023MathPrompterMR, Zhao2023AutomaticMS} encourage LLMs to generate diverse reasoning paths in different forms simultaneously, including text (CoT), code (PAL), and symbols (Equation) for a higher confidence answer. Automatic-CoT, Complexity-CoT, Synthetic Prompt and Boosted Prompt~\citep{Zhang2022AutomaticCO, Fu2022ComplexityBasedPF, Shao2023SyntheticPG, Pitis2023BoostedPE} enhance reasoning performance by optimizing the selection of demonstrations within the prompt. Tree-of-thought and Self-Evaluation~\citep{Yao2023TreeOT, Xie2023DecompositionER} extend the CoT into a search tree, obtaining more accurate answers through self-evaluation. 

\textbf{Finetuning-based Method} Another domain of study involves methods based on finetuning. These approaches involve finetuning open-source models, such as LLaMA, by incorporating insights from sophisticated closed-source LLMs. The fine-tuning approaches\citep{Yuan2023ScalingRO, Luo2023WizardMathEM, Yue2023MAmmoTHBM, Wang2023MakingLL, Yu2023MetaMathBY, Gou2023ToRAAT} also have the potential to improve the mathematical reasoning capabilities of LLMs. The essence of fine-tuning is centered around the development of quality datasets comprising question-response pairs.  Additionally, process-supervised training methods \citet{Lightman2023LetsVS, Wang2023MathShepherdVA} can also enhance the reasoning abilities of the LLMs.

\end{document}